\documentclass[11pt]{article}

\usepackage{microtype} 
\usepackage{booktabs}  
\usepackage{url}  

\usepackage{amsmath}
\usepackage{amsthm}
\usepackage{amssymb}

\newcommand{\argmin}{\arg\!\min}

\usepackage{todonotes}

\usepackage[finalworkshop]{automl}

\usepackage{natbib}
\usepackage{graphicx}
\usepackage{rotating}
\usepackage{listings}
\usepackage{booktabs}
\usepackage[capitalize]{cleveref}
\usepackage{wrapfig}
\usepackage{float}
\usepackage{enumitem}
\usepackage{todonotes}
\usepackage[symbol]{footmisc}

\title{Obeying the Order: Introducing \\ Ordered Transfer Hyperparameter Optimisation}

\author[1]{\nameemail{Sigrid Passano Hellan}{s.p.hellan@ed.ac.uk}\footnote[2]{Work done during an internship at AWS AI Labs.}}
\author[2]{\nameemail{Huibin Shen}{huibishe@amazon.com}\footnote[1]{Correspondence to: Huibin Shen <huibishe@amazon.com>.}}
\author[2]{\nameemail{Fran\c{c}ois-Xavier Aubet}{fxa@deepmind.com}}
\author[2]{\nameemail{David Salinas}{dsalina@amazon.com}}
\author[2]{\nameemail{Aaron Klein}{kleiaaro@amazon.com}}

\affil[1]{University of Edinburgh}

\affil[2]{AWS AI Labs}

\hypersetup{%
  pdfauthor={Sigrid Passano Hellan, Huibin Shen, Fran\c{c}ois-Xavier Aubet, David Salinas and Aaron Klein}, 
  pdftitle={Obeying the Order: Introducing Ordered Transfer Hyperparameter Optimisation},
  pdfsubject={We consider the problem of regularly tuning an ML system in a production system and show that state-of-the-art HPO transfer learning can be easily outperformed by taking the order into account. 
  },
  pdfkeywords={Hyperparameter optimisation, Transfer Learning, Bayesian Optimisation, Continual Learning, HPO deployment}
}

\newcommand{\randomsearch}{RandomSearch}
\newcommand{\bayesopt}{BO}
\newcommand{\boundingbox}{BoundingBox}
\newcommand{\zeroshot}{ZeroShot}
\newcommand{\quantiles}{CTS}
\newcommand{\studentbo}{SimpleOrdered}
\newcommand{\studentboshuffled}{SimpleOrderedShuffled}
\newcommand{\botransfer}{TransferBO}
\newcommand{\prevbo}{SimplePrevious}
\newcommand{\prevnobo}{SimplePreviousNoBO}
\newcommand{\dreordered}{DREOrdered}

\newcommand{\othpo}{OTHPO}
\newcommand{\othpolong}{ordered transfer HPO}

\newcommand{\othpoverylong}{ordered transfer hyperparameter optimisation}
\newcommand{\stantrans}{standard transfer HPO}

\begin{document}

\maketitle

\begin{abstract}
We introduce \textit{\othpoverylong{}} (\othpo{}), a version of transfer learning for hyperparameter optimisation (HPO) where the tasks follow a sequential order. Unlike for state-of-the-art transfer HPO, the assumption is that each task is most correlated to those immediately before it.
This matches many deployed settings, where hyperparameters are retuned as more data is collected; for instance tuning a sequence of movie recommendation systems as more movies and ratings are added.  
We propose a formal definition, outline the differences to related problems and propose a basic \othpo{} method that outperforms state-of-the-art transfer HPO.
We empirically show the importance of taking order into account using ten benchmarks.
The benchmarks are in the setting of gradually accumulating data, and span XGBoost, random forest, approximate k-nearest neighbor, elastic net, support vector machines and a separate real-world motivated optimisation problem. 
We open source the benchmarks to foster future research on \othpolong{}.

\end{abstract}

\section{Introduction}

All modern machine learning (ML) pipelines contain many hyperparameters that are critical for final performance, as they govern key parts of the training such as the optimisation (learning rate), the capacity of the model (number of layers or regularisation weights) or data augmentation. Hyperparameter optimisation (HPO) --- see e.g. the recent book by ~\cite{feurer2019hyperparameter} --- aims to find the optimal hyperparameters of a machine learning method by casting it as an optimisation problem: for each iteration a new set of hyperparameters is used to train and validate the method. 

\begin{figure}[ht]
    \centering
    \includegraphics[width=\linewidth]{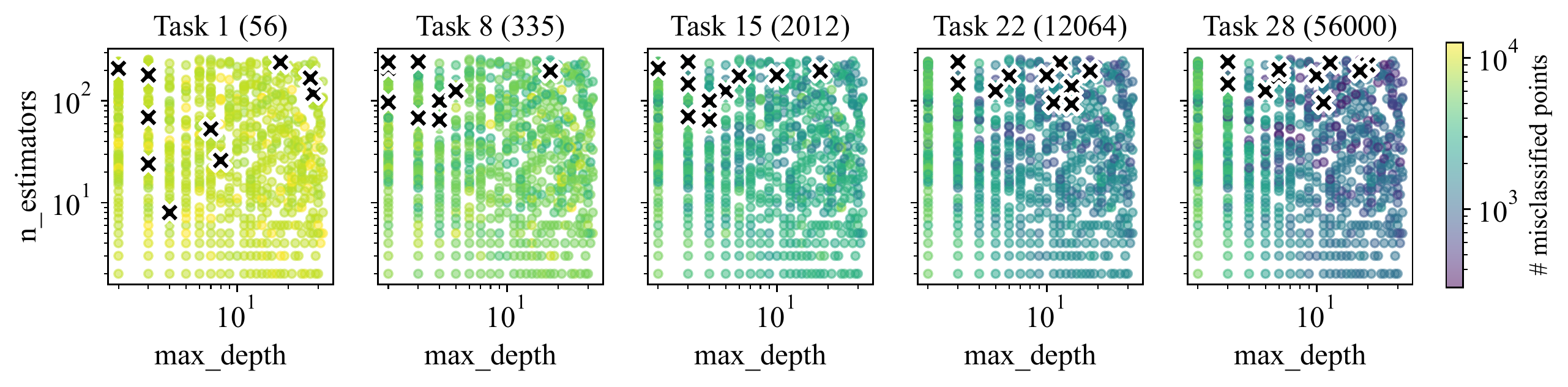}
    \caption{Evolution of the optimal XGBoost hyperparameters on MNIST for gradually increasing training set sizes (from 56 to 56000 points). 
    The black crosses indicate the 10 best hyperparameters, and can be seen to shift upwards to more estimators as more data is added. 
    }
  \label{xgboost-landscape}
\end{figure}

In practical scenarios, hyperparameters are not tuned once but many times. Consider a movie recommendation ML system being deployed. The model hyperparameters must be tuned frequently, given that the data set consistently evolves with new movies and users, which increases the data set size and gives a continuous shift in the optimal hyperparameter values. 
In particular, we expect some hyperparameters that define the regularisation or the capacity of the model to change as more data is observed. Smaller models might be initially superior for little data, as only simple rules can be learned, but, as more data becomes available, more expressive models start to become competitive, as they are able to identify smaller differences between the inputs. This point is illustrated in \cref{xgboost-landscape} which plots the validation performance of XGBoost with respect to the number of estimators and the maximum depth of each tree. As the data set size increases (from left to right) the optimal hyperparameter values change smoothly and more expressive models (i.e more and deeper trees) become superior. We call each hyperparameter optimisation of a model a \textit{task}.
 
\begin{figure}[h]
    \centering
    \includegraphics[width=0.5\linewidth]{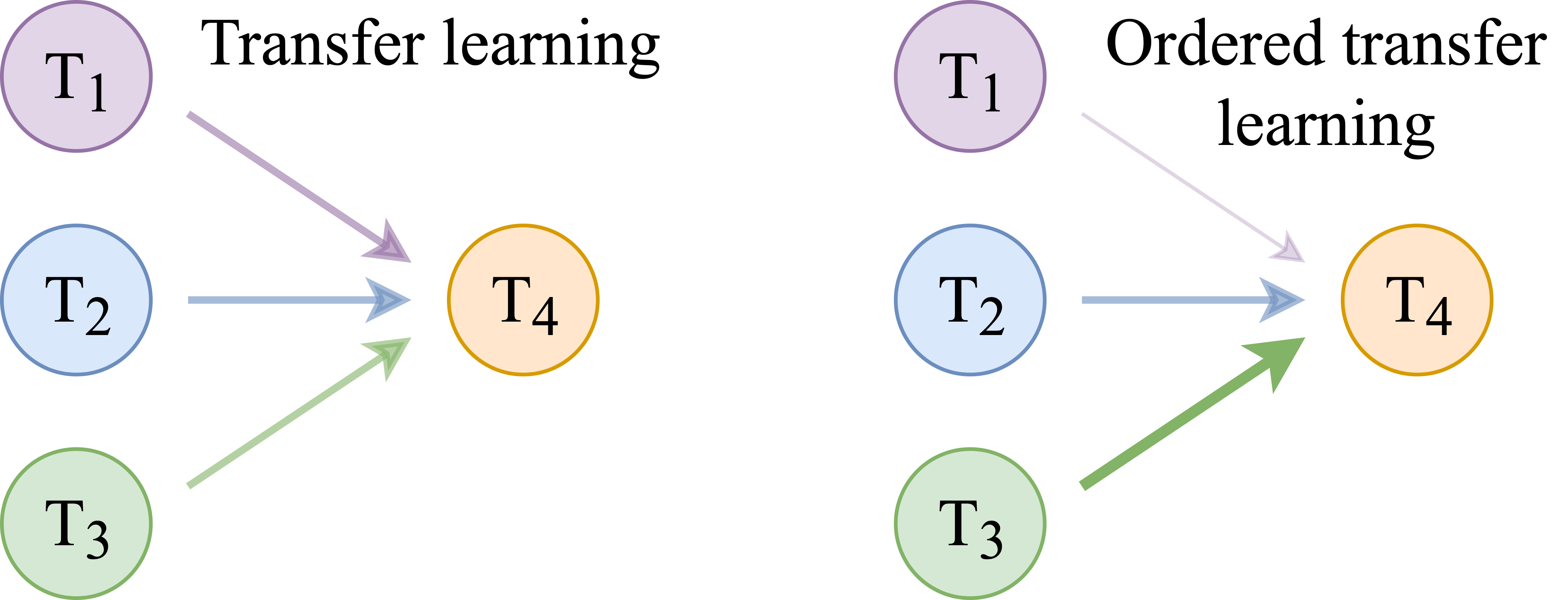}
    \caption{Contrasting \stantrans{} and \othpo{}. 
    In the former (left), when learning the next task, T\textsubscript{4}, each previous task (T\textsubscript{1}, T\textsubscript{2}, T\textsubscript{3}) is either assumed equally important or weighted according to meta-features or hyperparameter rank matching.
    For \othpo{} (right), more recent tasks are assumed more relevant, as illustrated by the difference in arrow widths.}
  \label{diagram-ordered-transfer}
\end{figure}

A popular family of approaches that can exploit information from previous tasks is transfer HPO~\citep{wistuba2015sequential,feurer2018practical,salinas2020quantile}. Such methods exploit data collected from the HPO of previous tasks to warm-start the optimisation on the current task. However, transfer HPO methods treat tasks as a set and ignore any intrinsic order. In practice, data is often collected in a sequence, for instance when a production system is tuned at frequent intervals.
We introduce \textit{\othpolong{}} (\othpo{}), as a special case of transfer HPO that exploits this sequential nature of tasks, enabling a better transfer of knowledge across tasks. See \cref{diagram-ordered-transfer} for an illustration. 
Our contributions are:
\begin{itemize}[noitemsep,nolistsep]
    \item We propose a formal definition of ordered transfer HPO and outline the differences to related problems, such as \stantrans{} and continual learning.  
    \item We provide ten benchmarks for this setup, integrated in the recently open-sourced HPO library Syne Tune \citep{salinas2022syne} to compare existing approaches for transfer HPO and foster the development of future methods.
    These benchmarks include XGBoost \citep{Chen_2016_XGBoost}, support vector machines, approximate k-nearest neighbor \citep{malkov2018efficient}, random forest \citep{wright2017ranger} and elastic net \citep{friedman2009glmnet} on various data sets, as well as a blackbox optimisation task based on SimOpt \citep{Eckman2023SimOptAT}.
    \item Our results in the setting of accumulating training data over time suggest that, in this setting, \othpo{} methods taking order into account are simple and performant, which provides  guidance for HPO practitioners in a deployed system.
\end{itemize}

\section{Related work}

\othpo{} is related to but distinct from \stantrans{} \citep{bai2023transfer}, continual learning \citep{van2019three,chaudhry2019tiny} and multi-fidelity HPO \citep{jamieson2016non,li2017hyperband}. 

\textbf{Standard transfer HPO} \citep{perrone2019learning,salinas2020quantile,wistuba2015sequential,horvath2021hyperparameter} typically transfers knowledge between data sets, and, compared to \othpo{} tasks, does not have an inherent ordering that one can exploit.
For example, \citet{wistuba2015sequential} used hyperparameter configurations evaluated on previous tasks to create a portfolio of well-performing configurations, which are sequentially evaluated on a new task.
\citet{perrone2019learning} reduced the search space for a new task by defining a bounding box around the best performing configurations on all previous tasks. 
To account for the variability in objective scales of different tasks, \citet{salinas2020quantile} learned a semi-parametric Gaussian Copula distribution across tasks. \citet{springenberg-nips16} used a Bayesian neural network to model the correlation between tasks for multi-task Bayesian optimisation. In a similar vein, \citet{perrone2019learning} trained neural networks to learn basis functions across tasks and combined this with Bayesian linear regression to obtain reliable uncertainty estimates for Bayesian optimisation. This idea was extended by \citet{horvath2021hyperparameter}, which regularised the basis functions to account for the changing complexity during the optimisation process.

In \cite{wistuba2021few}, the authors considered HPO as a few-shot learning problem where a Deep GP (Gaussian process) model was trained jointly on a set of meta-tasks by few-shot learning. For a target task they then started from the initialised kernel parameters, before fine-tuning the model with a few hyperparameter evaluations.

Some transfer learning HPO methods such as \cite{wistuba2015learning} proposed leveraging meta-features of previous tasks to exploit task similarities. However, those methods rely on manual engineering of task features which are critical for final performance. To tackle this issue, \cite{jomaa2021transfer} proposed using Deep GPs to learn meta-features in an end-to-end fashion, which shows encouraging results in combating negative transfers. In what follows, we use \stantrans{} as a shorthand for non-ordered transfer learning. 

Both \othpo{} and \textbf{continual learning} consider sequences of tasks. 
But continual learning is concerned with maintaining model performance on previous tasks, typically keeping the hyperparameters constant. We, on the other hand, want to optimise the hyperparameters for the current task and do not need the new model to perform well on the previous tasks. 

In contrast to \textbf{multi-fidelity HPO}, \othpo{} cares about the performance at each level. While a subset of the training data could be used in the multi-fidelity setting as a heuristic for later performance \citep{li2017hyperband, klein2017fast}, we consider the performance on the earlier task as a goal in itself. The idea \citep{zappella-arxiv21} of multi-fidelity HPO has been also extended to the \stantrans{} setting.

Previous work has also been motivated by the idea of considering HPO of a model under change as a sequence of tasks \citep{golovin2017google,zhang2019lifelong,stoll2020}, 
but to our knowledge none have explicitly evaluated the importance of the ordering and they exhibit a few notable differences to our work. 
In \cite{stoll2020}, the ordering is implicitly used by only transferring from the latest task with a different search space and underlying algorithm. \cite{golovin2017google} motivate their work through HPO on different tasks in a sequence, but do not present results on HPO or compare to \stantrans{}. 
The open source version of the software does not include transfer learning, so is not included as a baseline in this paper.

\cite{zhang2019lifelong} is most related to our work as they identify the practical problem of slowly evolving data sets and the need to perform transfer HPO. 
However, they do not explicitly compare using the task ordering to \stantrans{}.
And while their evaluation considers the best possible performance on a new task, we show the large potential speed-ups possible by using \othpo{}, since we show improved results after only one hyperparameter evaluation.
The \othpo{} method we propose is much simpler than those in \cite{golovin2017google,zhang2019lifelong}. This means we can directly evaluate the benefit of taking the ordering into account. 

Before going further, we want to remind our readers that our goal is not to propose another transfer HPO method but rather to introduce a setting that is relevant for practitioners using HPO in a deployed system, and demonstrate the potential of utilising the sequential nature of the tasks.

\section{Problem definition} \label{sec-problem-definition}

\newcommand{\cont}{\boldsymbol{\phi}}
\newcommand{\eval}{\boldsymbol{y}}
\newcommand{\hyps}{\boldsymbol{\theta}}
\newcommand{\hypspace}{\Theta}
\newcommand{\contspace}{\mathbb{V}}

\newcommand{\probdefhpo}{\displaystyle \hyps^*~=~\argmin_{\hyps \in \hypspace}~f(\hyps)}

\newcommand{\probdef}{\displaystyle \hyps^*_{i}~=~\argmin_{\hyps}~f(\hyps;\cont_i)}
\newcommand{\funcdef}{f:~\hypspace~\times \contspace~\rightarrow~\mathbb{R}}

\begin{figure}[h]
    \centering
    \includegraphics[width=0.82\linewidth]{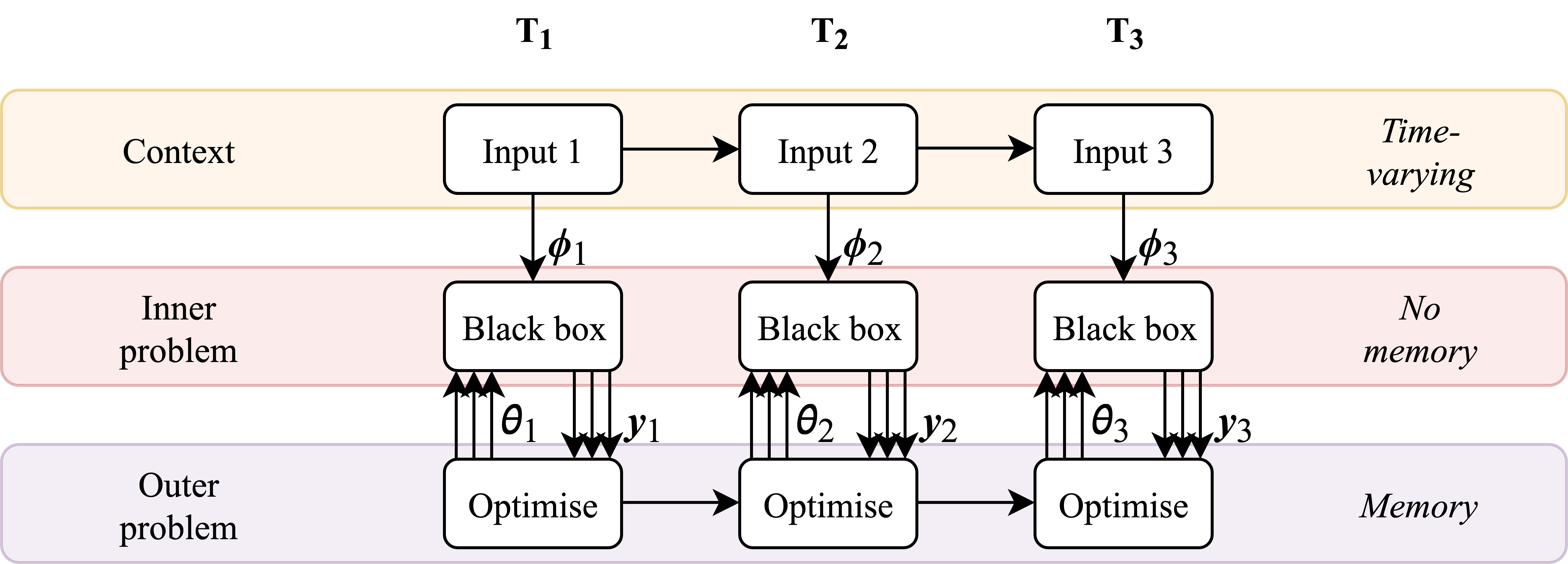}
    \caption{Structure of an \othpolong{} problem. The tasks T\textsubscript{1}, T\textsubscript{2}, T\textsubscript{3} have an inherent order in the outer problem stemming from the drift in the context. There is no connection between the inner problems. For the notation see \cref{sec-problem-definition}.}
  \label{diagram-problem-structure}
\end{figure}

Let  $f: \hypspace \rightarrow \mathbb{R}$ denote the validation performance of a machine learning algorithm after training with hyperparameters $\hyps \in \hypspace$. 
HPO treats the search for the optimal hyperparameters as a global optimisation problem $\probdefhpo$. The space of all possible  hyperparameter configurations $\hyps \in \hypspace$ is called the configuration space. Due to the intrinsic randomness of most machine learning methods, for example random weight initialisation or mini-batch sampling, we observe $f$ only with noise:  $\eval = f(\hyps) + \epsilon$, where $\epsilon \sim \mathcal{N}(0, \sigma^2)$.

In practice, we often face the same HPO problem repeatedly on different tasks, where the configuration space and the underlying machine learning algorithm are the same, but training and validation data sets change.
To share knowledge across HPO tasks, we treat the objective functions as a series of related global optimisation problems, see \cref{diagram-problem-structure}.
More formally, we augment the definition of our objective function $\funcdef$ by another input $\cont_i \in \contspace$ that denotes the current task $i$. Now, \othpo{} assumes that tasks come in a sequence, such that task $i$ is more similar to task $i-1$ than to task $i-2$.

For task $i$, we collect evaluations $\eval_{i,m}=f(\hyps_{i,m};\cont_i) + \epsilon$ through the optimiser in the outer problem. The subscript $m$ denotes which of the $M_i$ evaluations of the inner problem $i$ is given. When deciding what configuration to try next, the outer optimiser has access to the evaluations of previous tasks 
$\{\hyps_{j,m}, \eval_{j, m}\}_{j = 1}^{i-1}, _{m = 1}^{M_j}$ and all the $N$ finished evaluations in the current task $\{\hyps_{i,m}, \eval_{i, m}\}_{m = 1}^{N}$.
We assume that the search space $\hypspace$ remains the same between tasks.

\newcommand{\simoptbench}{NewsVendor}
\newcommand{\svmfirst}{SVM 1220}

\section{Benchmarks}
\label{sec:bench}
We propose 10 benchmarks to evaluate methods on \othpo{}, summarised in \cref{tab:relating-benchmarks}.
We implement our benchmarks in Syne Tune \citep{salinas2022syne}, making them easily available to everyone. 
In this section we describe each benchmark in more depth.

\begin{table}[h]
\small
\centering
\begin{tabular}{@{}llll@{}}
\toprule
                & XGBoost                 & YAHPO             & \simoptbench{}   \\ \midrule
Context         & Training data size     & Training data size & Environment settings   \\
Inner problem   & Minimise error            & Maximise AUC      & Simulate profit  \\
Outer problem   & HPO                    & HPO                & Parameter optimisation         \\
Number of tasks & 28                     & 20                 & 9                       \\
Number of benchmarks & 1                     & 8                 & 1                       \\ \bottomrule
\end{tabular}
\caption{Overview of benchmarks, using concepts from \cref{diagram-problem-structure}.}
\label{tab:relating-benchmarks}
\end{table}

\paragraph{XGBoost on MNIST} In a deployed setting, one collects more training data as time passes. When refitting the model on more data it is likely that the optimal hyperparameters on earlier data sets are not optimal anymore. We propose an evaluation benchmark for this setting by training an XGBoost classifier \citep{Chen_2016_XGBoost} on the MNIST data set \citep{OpenML2013} with increasing training set sizes.
We tune four hyperparameters: \lstinline{max_depth}, \lstinline{n_estimators}, \lstinline{min_child_weight} and \lstinline{learning_rate}.
We have 28 tasks, with training set sizes regularly selected in the log space, ranging from 56 to 56000 training examples (see \cref{app-xgboost-additional}). Some of these tasks are shown in \cref{xgboost-landscape}. We use surrogate models fit on 1000 hyperparameter evaluations to avoid any model training during HPO.
We ensure all ten classes are represented in every task.
Our optimisation metric is the number of misclassified points in a validation set comprising\textbf{} 14000 examples.

\begin{figure}[h]
    \centering
    \begin{subfigure}[b]{\linewidth}
         \centering
    \includegraphics[width=\linewidth]{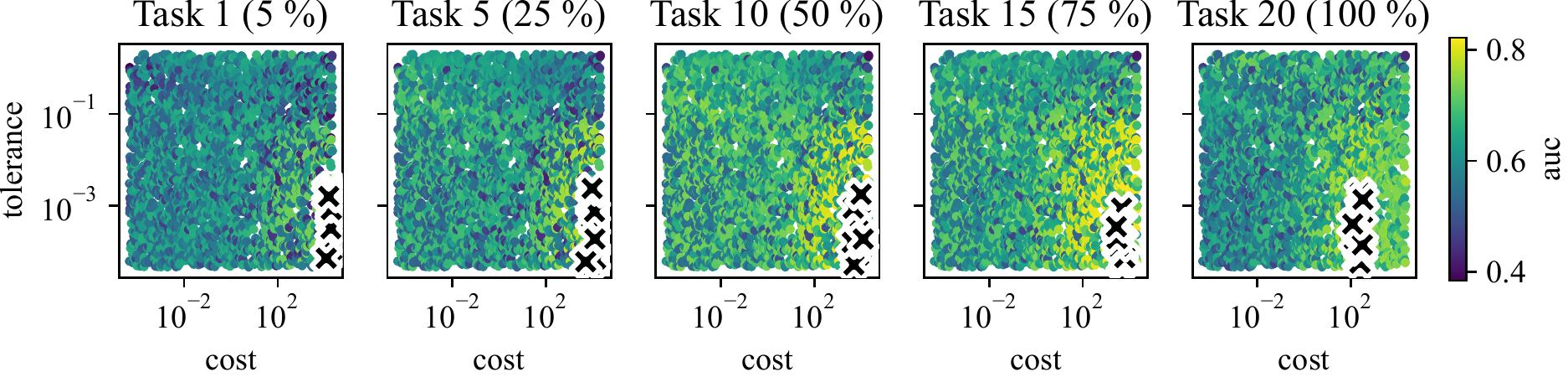}
  \end{subfigure} 
\begin{subfigure}[b]{\linewidth}
    \centering
    \includegraphics[width=\linewidth]{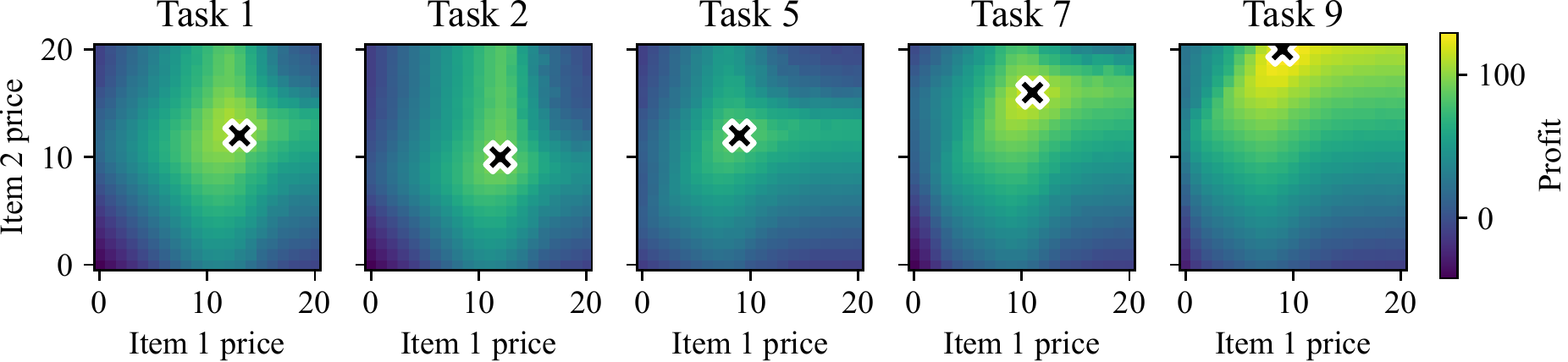}
    \end{subfigure} 
    \caption{Hyperparameter landscapes of YAHPO SVM 1220 (top) and \simoptbench{} (bottom). Crosses indicate the 10 best configurations per task for YAHPO and the best one for \simoptbench{}.
    }
    \label{fig:yahpo-simopt-landscapes}
\end{figure}

\paragraph{YAHPO} The next 8 benchmarks are drawn from YAHPO Gym \citep{pfisterer2022yahpo}, a recently published HPO benchmark suite containing a variety of HPO problems. We focus on a subset of scenarios from RandomRobot version 2 (rbv2) because they have different training set sizes available. Our initial investigation suggests that, depending on the ML model and the data set, the top performing hyperparameters are either smoothly changing over increasing training set size or stay in a similar region. We select smoothly changing ones for our benchmarks. An example is given in \cref{fig:yahpo-simopt-landscapes} (top), for more see \cref{additional-yahpo-plots}. In YAHPO, surrogate models are also used to predict hyperparameter performances for faster experimentation.

We consider four diverse ML models with two data sets each, so eight data sets in total, and optimise the AUC. The models considered \citep{binder2020collecting} are SVM (support vector machines), AKNN (approximate k-nearest neighbor, \cite{malkov2018efficient}), ranger (random forest, \cite{wright2017ranger}) and glmnet (elastic net, \cite{friedman2009glmnet}). The data set ID will be shown next to the algorithm, e.g. SVM 1220.
We create 20 tasks by gradually increasing the size of the training data set from 5\% to 100 \%.

\paragraph{\simoptbench{}} The benchmarks presented so far are focused on the change in the optimal hyperparameters of ML models as  data set sizes increase. However, there are many other kinds of systems which need to be optimised periodically in evolving environmental conditions. The \simoptbench{} benchmark adds one such system to our set of benchmarks. 
In \simoptbench{}, the aim is to maximise profit by setting the prices for three item categories given the current uncertain demand for the items.
This benchmark is based on the Dynamic News problem from the SimOpt library that contains many simulation-optimisation problems and solvers \citep{Eckman2023SimOptAT}.
Over time the item demand, or utility, is influenced by external factors and evolves. Here we simulate the change by following a random walk on the utility,
resulting in the sequence of tasks. Note that this means the context is not necessarily changing in a single direction as for XGBoost and YAHPO.
This is illustrated in \cref{fig:yahpo-simopt-landscapes} (bottom). 
We consider a sequence of 9 tasks, with three item categories, making the outer problem 3-dimensional with an integer search space limited between 0 and 20.

\section{Experiments} \label{sec-experiments}

To show the potential gain available in transfer HPO from taking the task order into account, we compare simple \othpo{} methods to non-transfer Bayesian optimisation and \stantrans{} methods.
Our code is available at \url{https://github.com/sighellan/syne-tune/tree/othpo-results}.

\subsection{Baselines} \label{sec-baselines}

We consider several non-transfer and transfer HPO methods from the literature as baselines:
\begin{itemize}[noitemsep]
    \item \textbf{\randomsearch{}}: Sample configurations uniformly at random from the search space.
    \item \textbf{\bayesopt{}} \citep{snoek2012practical}: Run Bayesian optimisation with no transfer between tasks.
    \item \textbf{\boundingbox{}} \citep{perrone2019learning}: Shrink the search space of \bayesopt{} to the bounding box of optima on previous tasks. Note that this means the search space cannot increase for future tasks and it requires two finished tasks so a box can be computed. 
    \item \textbf{\zeroshot{}} \citep{wistuba2015sequential}: Learn a portfolio of complementary hyperparameters with greedy selection based on previous tasks performance. The configurations of the portfolio are then evaluated sequentially.   
    \item \textbf{\quantiles{}} (Copula Thompson Sampling, \cite{salinas2020quantile}): 
     Map the evaluations to quantiles within each task and learn a probabilistic model to predict the quantiles. 
     For a new task, the method then samples the performance of each candidate configuration and picks the configuration with the lowest sampled value.
\end{itemize}

We use implementations in Syne Tune for our baselines and BoTorch \citep{balandat2020botorch} for \bayesopt{} and the transfer learning methods relying on BO. We use a Mat\'ern 5/2 kernel and a Monte Carlo version of expected improvement \citep{jones1998efficient}, see \cref{app-bo-details}.
The HPO tasks are sequentially evaluated, with the evaluations collected in one task available for the subsequent tasks.

\subsection{Simple \othpolong{} methods} 
\label{sec-ordered-methods}
\begin{itemize}[noitemsep]
    \item \textbf{\botransfer{}}: 
    \textit{Extend the BO surrogate model to also take the task order as an input feature.}
    All the evaluations from previous tasks are used to train a GP where the task order is explicitly modelled through a task feature: For \simoptbench{}, we use the task index as the feature; For the other benchmarks, we use the training set size. When the tasks are close in the task feature, they have more impact on each other than the tasks that are distant. The idea of modelling training set sizes in the surrogate model have been widely used \citep{klein2017fast, klein2020model}. We use the simplest form where the same kernel function is applied on both hyperparameters and training set sizes.
    \item \textbf{\studentbo{}}: \textit{Standard BO, but start by evaluating top-performing hyperparameters from the previous tasks.}
    For a new task, the first $N$ hyperparameter configurations come from the top hyperparameter configurations of each of the previous $N$ tasks, starting with the most recent one and continuing in reverse order of time. 
    If there are not $N$ previous tasks yet, we continue by taking second-optimal points from those available and so on. If there are repeated optimal configurations we skip to earlier tasks.
    Then we continue with standard BO.
    We set $N=5$ in our experiments. We also test a variant \textbf{\studentboshuffled{}} in ablations where we still take top configuration from previous tasks, but in random order.
    \item \textbf{\prevbo{}}: \textit{Same as \studentbo{}, but only using the last previous task.} The initial $N=5$ hyperparameters are the best $N$ hyperparameters from the previous task. It is a simplified version of \cite{feurer2015initializing} where meta-feature computation can be avoided due to our assumption that the closest task in the sequence is the most similar task. We also test a variant \textbf{\prevnobo{}} which uses all the hyperparameters from the previous task, sorted by decreasing performance, without any BO.
\end{itemize}

\subsection{Experimental setup and metric}

\textbf{Experimental setup:} 
For each benchmark described in \cref{sec:bench}, we sequentially apply HPO and transfer HPO methods for each tuning task with the number of hyperparameter evaluations restricted to 25. We rerun each experiment with 50 seeds and report average performance $\pm$ 2 standard errors when plotting method results.
The transfer learning methods assume evaluations from at least one previous task. All these methods therefore use \bayesopt{} to collect evaluations on the first task.
\boundingbox{} also uses it on the second task, see \cref{app-bo-start}.

\textbf{Metric:}
For ease of aggregation, we use the Normalised Score from \citep[eq. 3]{cowen2022hebo}. Let $i$ index tasks, $m$ index iterations within a HPO task and $M$ be the maximum number of hyperparameter evaluations for each task. The score is defined as
$ \displaystyle 100 * ( L_{i,m} - L_{i,M}^{\mathrm{best}}) /( L_{i,M}^{\mathrm{RS}} - L_{i,M}^{\mathrm{best}} ) $,
where $L_{i,m}$ is the mean loss across replications for a method on task $i$ at iteration $m$, $L_{i,M}^{\mathrm{best}}$ is the estimated best solution for the task and $L_{i,M}^{\mathrm{RS}}$ is the mean performance of \randomsearch{} on the task by the final iteration. 
For $L_{i,M}^{\mathrm{best}}$ we use the best mean obtained across the compared methods.  

\begin{figure}[h]
    \centering
    \includegraphics[width=0.85\linewidth]{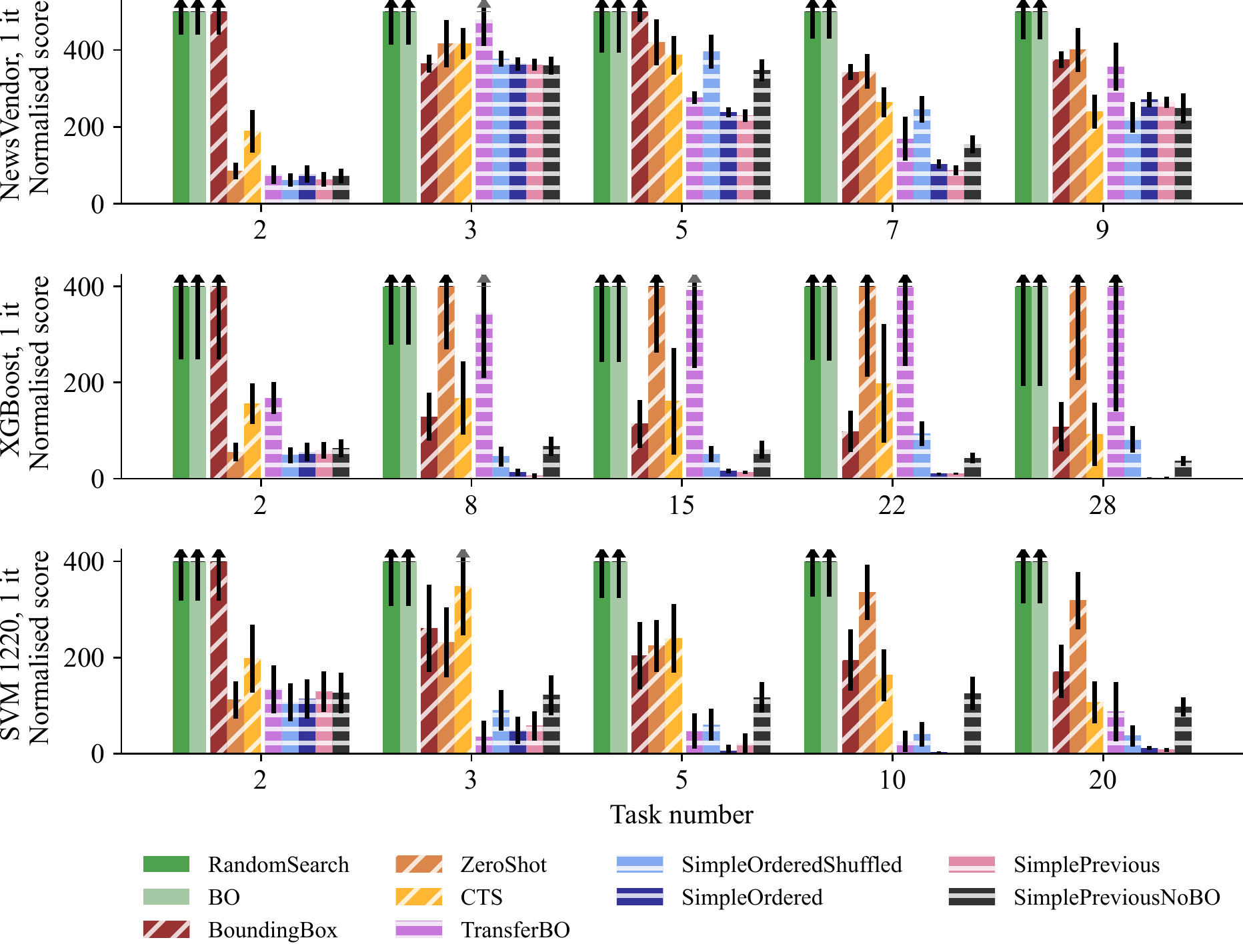}
    \caption{Mean normalised scores after first configuration (lower is better). \simoptbench{} (top), XGBoost (middle) and \svmfirst{} (bottom).  Black arrows indicate that the mean was above the plotted range, grey arrows that the standard error range was above. The methods are grouped by colour with \othpo{} methods in blue, purple and pink, \stantrans{} methods in yellow and red, and methods with no transfer in green. Task 2 is the first transfer task.}
  \label{fig:all-normalised-task}
\end{figure}

\begin{figure}[h]
    \centering
    \includegraphics[width=0.7\linewidth]{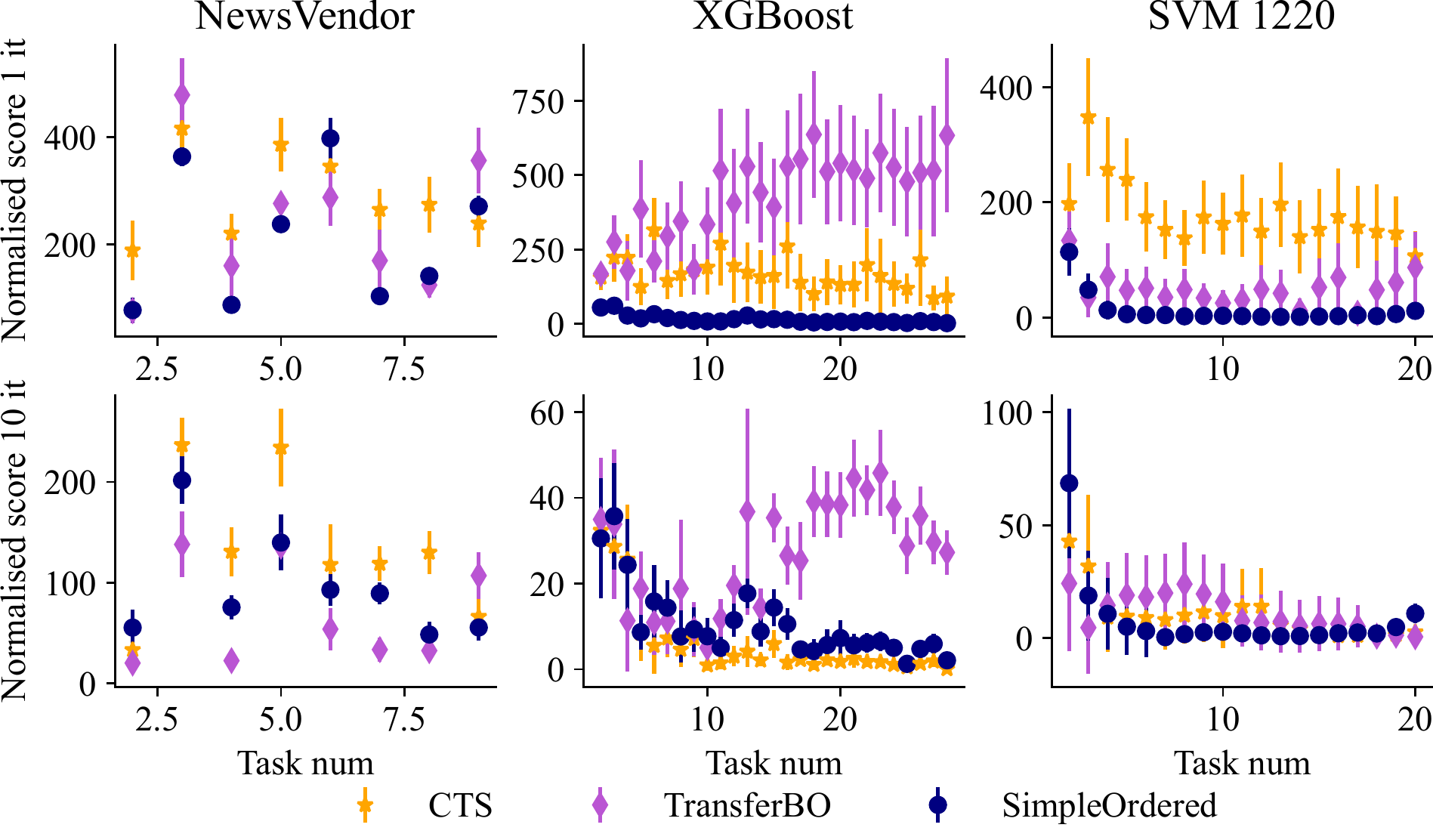}
    \caption{Mean normalised scores after first (top) and tenth (bottom) configuration. Top: \studentbo{} clearly outperforms \botransfer{} and \quantiles{}. Bottom: For \simoptbench{} and XGBoost we see that \botransfer{} and \quantiles{}, respectively, are able to find better solutions than \studentbo{}. 
    }
  \label{fig:compare-studentBO-shuffled}
\end{figure}

\begin{figure}[h]
    \centering
    \includegraphics[width=0.7\linewidth]{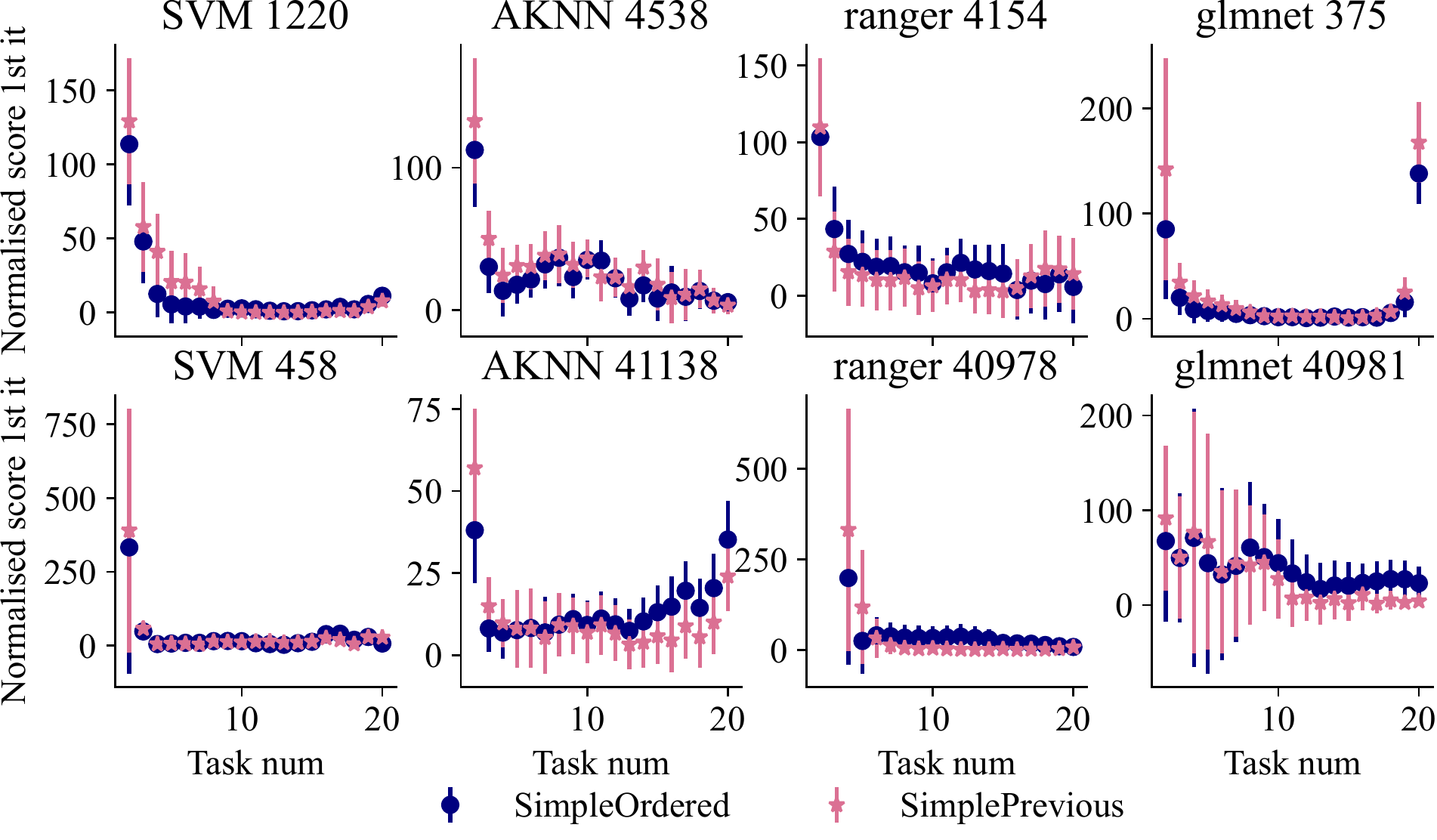}
    \caption{Mean normalised scores after first configuration evaluation for all eight YAHPO combinations considered.  We see that for \studentbo{} and \prevbo{} perform very similarly. }
  \label{fig:yahpo-additional-scenarios}
\end{figure}

The Normalised Score computes the loss distance at an iteration to the best solution, normalised by the loss distance between \randomsearch{} at the final iteration $M$ and the best solution, thus the smaller the better, ideally 0.
It allows easy comparison between minimisation (XGBoost) and maximisation (\simoptbench{}, YAHPO) benchmarks, and also normalises tasks by difficulty and away from the scale of the optimisation metric. 
We present results after 1 and 10 iterations. \cref{fig:all-normalised-task} compares all ten methods on \simoptbench{}, XGBoost and \svmfirst{} after 1 iteration. \cref{fig:yahpo-additional-scenarios} compares the best \othpo{} methods, \studentbo{} and \prevbo{}, on the eight YAHPO combinations after 1 iteration. \cref{fig:compare-studentBO-shuffled} compares \studentbo{} and \botransfer{} to the top-performing \stantrans{} method \quantiles{} after 1 and 10 iterations. 
Further results are given in \cref{app-additional-results}.

\subsection{First evaluation: \studentbo{} and \prevbo{} beat \stantrans{}}

Using ordering gives a large benefit at the first evaluation. This can be seen in \cref{fig:all-normalised-task} and \cref{fig:compare-studentBO-shuffled} (top). 
We first note that all the transfer HPO methods, including ours, outperform non-transfer methods. \studentbo{} mostly beats --- or at least is on par compared to --- all the baselines. 
For \simoptbench{}, \studentbo{} does slightly worse than for the other benchmarks. This might be because the number of previous tasks that can be used is relatively small, and the changes between tasks is not a gradual shift in one direction like in the other benchmarks. We note that \studentbo{} is a very simple method in comparison to the \stantrans{} baseline methods. \botransfer{} performs worse than \studentbo{}. This shows that \botransfer{} is not able to use the task index effectively.
We investigate the sampling pattern in \cref{app-sampling-locations}.

In \cref{fig:all-normalised-task} we see that \studentbo{} and \prevbo{} outperform \stantrans{}.
We compare \studentbo{} and \prevbo{} in \cref{fig:yahpo-additional-scenarios},
and see that the performance is very similar, with \studentbo{} slightly preferable on early tasks and \prevbo{} on later tasks. 
This highlights the large impact of the most recent task.

We also further analyse the impact of ordering and the surrogate in our ablations in \cref{app-ablations}, which shows that the lack of ordering in \studentboshuffled{} is detrimental, as is the lack of \bayesopt{} in \prevnobo{}.

\subsection{Ordered advantage reduces with more evaluations} \label{sec-othpo-adv-reduces}

While \studentbo{} and \prevbo{} are clearly better after the first evaluation, this becomes benchmark-specific after the fifth evaluation. As can be seen in \cref{fig:compare-studentBO-shuffled} (bottom), \botransfer{} is best for \simoptbench{}, \quantiles{} is best for XGBoost and \studentbo{} is best for \svmfirst{}.
We investigate this further in \cref{app-rankings,app-additional-bars,app-additional-iteration-curves}. It is difficult to declare a method best because the variance between runs and between benchmarks becomes too high.
The general trend is that \studentbo{} and \prevbo{} pick better first configurations, and the other methods catch up  with more evaluations.
But \studentbo{} and \prevbo{} remain reliable choices even for greater number of evaluations. While \botransfer{} and \quantiles{} beat them on individual benchmarks, they do worse on other ones.
From this we conclude that; a) there is no universal method to prefer after five evaluations, it depends on the set of tasks; and b) there is scope for improved methods combining \studentbo{} with either \botransfer{} or \quantiles{} to come up with a stronger method. We also expect other more sophisticated \othpo{} methods to be able to outperform \studentbo{}.

\subsection{How much better will the trained models be?} \label{res-downstream}

The normalised score is very useful for comparing methods across tasks and benchmarks. But it abstracts away the potential performance gain, leaving the question: \textit{how much better will my model perform after the first configuration if I use \studentbo{} instead of \quantiles{}?}

There are two benefits: lower variance and improved mean. 
Using the metrics of the underlying tasks, we get a mean improvement of 21.7\% in profit for \simoptbench{}, 22.5 \% in number of misclassified points in XGBoost and 5.8 \% in AUC for \svmfirst{}.
The lower variance is also very valuable, as it reduces the need to retrain models with multiple seeds to get a good model.
The reduction in standard error of using \studentbo{} instead of \quantiles{} is 61.3 \% for \simoptbench{}, 92.5 \% for XGBoost and 89.4 \% for \svmfirst{}.
We summarise these numbers in a table in \cref{app-downstream}, where we also give error bounds.
The higher variance of \quantiles{} is also visible in \cref{fig:all-normalised-task,fig:compare-studentBO-shuffled}.

\section{Conclusion}

We introduced the problem of ordered transfer learning, motivated by the need of tuning regularly deployed models over time. We proposed a novel set of benchmarks to evaluate the performance of HPO methods in this setting, containing a blackbox optimisation problem, as well as HPO of XGBoost, SVM, random forest, elastic net and approximate k-nearest neighbor. We illustrated the key difference with \stantrans{} approaches 
and showed how simple methods taking the order into account can outperform more sophisticated transfer methods by better tracking smooth shifts of the hyperparameter landscape. We hope that our simple methods will be useful to enable the regular tuning of deployed methods while containing tuning costs, and that the benchmarks will enable evaluating methods in this practical setting.

\textbf{Practical recommendation:} 
Our results show that in this setting of accumulating data, \studentbo{} performs well, especially on early iterations for a new task. We therefore recommend practitioners to start with this simple method before trying more sophisticated ones.

\section{Limitations} \label{sec-limitations}

We focused on a situation that commonly occurs for deployed models, namely that the size of the training data increases over time, but other sequences of tasks should also be explored, e.g. when the model capacity or number of classes increases over time. In addition, while we show good performance for the simple method proposed, we believe further gain can be achieved with more sophisticated methods that decide for instance whether tuning is needed on the new tasks and automatically terminates in cases where tasks do not change, similarly to \cite{makarova22a}.

\section{Broader Impact Statement}

The intended impact is to make a subset of transfer HPO problems more efficient, with the positive impact of reducing energy consumption from training models. There is a risk that this will instead lead to the same computational budget being used and higher accuracies obtained, but that is just the lack of a benefit, not a harm in itself. However, we also show the benefit of doing hyperparameter optimisation for subsequent tasks as data set sizes increase. This could lead to more models being trained. By contributing simulation-based benchmarks the total energy consumption of future work should be reduced as the models do not need to be retrained. We see no ethical concerns with the data sets used.

\newpage
\section{Submission Checklist}

\begin{enumerate}
\item For all authors\dots
  \begin{enumerate}
  \item Do the main claims made in the abstract and introduction accurately
    reflect the paper's contributions and scope?
    \answerYes{[We also itemise our contributions in the introduction.]}
  \item Did you describe the limitations of your work?
    \answerYes{[See \cref{sec-limitations}.]}
  \item Did you discuss any potential negative societal impacts of your work?
    \answerYes{[See the Broader Impact Statement.]}
  \item Have you read the ethics author's and review guidelines and ensured that
    your paper conforms to them? \url{https://automl.cc/ethics-accessibility/}
    \answerYes{[The ethics guidelines at \url{https://2023.automl.cc/ethics/}.]}
  \end{enumerate}
\item If you are including theoretical results\dots
  \begin{enumerate}
  \item Did you state the full set of assumptions of all theoretical results?
    \answerNA{}
  \item Did you include complete proofs of all theoretical results?
    \answerNA{}
  \end{enumerate}
\item If you ran experiments\dots
  \begin{enumerate}
  \item Did you include the code, data, and instructions needed to reproduce the
    main experimental results, including all requirements (e.g.,
    \texttt{requirements.txt} with explicit version), an instructive
    \texttt{README} with installation, and execution commands (either in the
    supplemental material or as a \textsc{url})?
    \answerYes{[These are given at \url{https://github.com/sighellan/syne-tune/tree/othpo-results}, as stated in \cref{sec-experiments}. We provide requirement files for running the experiments both locally and remotely, and a README with step-by-step instructions.]}
  \item Did you include the raw results of running the given instructions on the
    given code and data?
    \answerYes{[These are available together with the rest of the code at \url{https://github.com/sighellan/syne-tune/tree/othpo-results}.]}
  \item Did you include scripts and commands that can be used to generate the
    figures and tables in your paper based on the raw results of the code, data,
    and instructions given?
    \answerYes{[For all plots containing data. Two of our figures (\cref{diagram-ordered-transfer,diagram-problem-structure}) are illustrative and do not include data; scripts are not included for these. In the README we provide a list of what scripts generate what figures.]}
  \item Did you ensure sufficient code quality such that your code can be safely
    executed and the code is properly documented?
    \answerYes{[See the README for more details.]}
  \item Did you specify all the training details (e.g., data splits,
    pre-processing, search spaces, fixed hyperparameter settings, and how they
    were chosen)?
    \answerYes{[This is also fully specified in the code to make it reproducible.]}
  \item Did you ensure that you compared different methods (including your own)
    exactly on the same benchmarks, including the same datasets, search space,
    code for training and hyperparameters for that code?
    \answerYes{[We collect the results for the different methods on all benchmarks ourselves, and keep the settings of the benchmarks constant.]}
  \item Did you run ablation studies to assess the impact of different
    components of your approach?
    \answerYes{[We provide results both for \studentbo{} and for a non-order version, \studentboshuffled{}. And for \prevbo{} and a version without BO, \prevnobo{}. See \cref{app-ablations}.]}
  \item Did you use the same evaluation protocol for the methods being compared?
    \answerYes{[This was automated in the file \lstinline{preprocess_results.py}.]}
  \item Did you compare performance over time?
    \answerYes{[We discuss this in \cref{sec-othpo-adv-reduces}, and present additional results in \cref{app-rankings,app-additional-bars,app-additional-iteration-curves}.
    ]}
  \item Did you perform multiple runs of your experiments and report random seeds?
    \answerYes{[We rerun each experiment 50 times. The seeds are given in the code and in \cref{app-bo-start}.]}
  \item Did you report error bars (e.g., with respect to the random seed after
    running experiments multiple times)?
    \answerYes{[Our plots contain error bars. The error range for the values in \cref{res-downstream} are given in \cref{app-downstream}. ]}
  \item Did you use tabular or surrogate benchmarks for in-depth evaluations?
    \answerYes{[We used the YAHPO surrogate benchmark, and used Syne Tune to generate a surrogate on top of our XGBoost evaluations.]}
  \item Did you include the total amount of compute and the type of resources
    used (e.g., type of \textsc{gpu}s, internal cluster, or cloud provider)?
    \answerYes{[This is given in \cref{app-compute-budget}.]}
  \item Did you report how you tuned hyperparameters, and what time and
    resources this required (if they were not automatically tuned by your AutoML
    method, e.g. in a \textsc{nas} approach; and also hyperparameters of your
    own method)?
    \answerNA{[We did not tune the hyperparameters of our methods. \studentbo{} and \studentboshuffled{} have a parameter $N$ which we set to 5 without tuning.]}
  \end{enumerate}
\item If you are using existing assets (e.g., code, data, models) or
  curating/releasing new assets\dots
  \begin{enumerate}
  \item If your work uses existing assets, did you cite the creators?
    \answerYes{[These are cited in the paper. They are also repeated in \cref{app-assets-used}.]}
  \item Did you mention the license of the assets?
    \answerYes{[We list the licenses of the assets used in \cref{app-assets-used}.]}
  \item Did you include any new assets either in the supplemental material or as
    a \textsc{url}?
    \answerYes{[We include all the code and results obtained as supplementary material.]}
  \item Did you discuss whether and how consent was obtained from people whose
    data you're using/curating?
    \answerNo{[The only data set we handle is MNIST. There are other data sets listed in \cref{app-assets-used}, but we do not use these directly -- they are listed for completeness. We only use the models of hyperparameter performance in YAHPO. These data sets were used to learn the published YAHPO models.]}
  \item Did you discuss whether the data you are using/curating contains
    personally identifiable information or offensive content?
    \answerYes{[This is discussed in \cref{app-assets-used}. It does not.]}
  \end{enumerate}
\item If you used crowdsourcing or conducted research with human subjects\dots
  \begin{enumerate}
  \item Did you include the full text of instructions given to participants and
    screenshots, if applicable?
    \answerNA{}
  \item Did you describe any potential participant risks, with links to
    Institutional Review Board (\textsc{irb}) approvals, if applicable?
    \answerNA{}
  \item Did you include the estimated hourly wage paid to participants and the
    total amount spent on participant compensation?
    \answerNA{}
  \end{enumerate}
\end{enumerate}

\begin{acknowledgements}

The authors would like to thank Jan Gasthaus, Valerio Perrone, Martin Wistuba and Michael Bohlke-Schneider for help with the project.

Hellan was supported by the EPSRC Centre for Doctoral Training in Data Science, funded by the UK Engineering and Physical Sciences Research Council (grant EP/L016427/1) and the University of Edinburgh.

\end{acknowledgements}


\bibliography{references}



\appendix

\section{Assets used} \label{app-assets-used}

We use benchmarks from three families:
\begin{itemize}[noitemsep,nolistsep]
    \item YAHPO \citep{pfisterer2022yahpo} (Apache License 2.0)
    \begin{itemize}[noitemsep,nolistsep]
        \item We only handle the surrogate benchmarks in YAHPO, not the data used to generate them. But we list these data sets here. For the data sets citing \cite{Dua:2019} the authors of the repository request citation. 
        \item Data set 1220 \citep{kdd2012}. Requires attribution, and the data are restricted to be used for scientific research purposes only. License: Public
        \item Data set 4538. Requires citation of \citep{Dua:2019} and a relevant paper, e.g. \citep{madeo2013gesture} License: Public
        \item Data set 4154. License: Public
        \item Data set 375 \citep{Dua:2019,kudo1999multidimensional}. Author asked to be informed about published work using the data. License: Public
        \item Data set 458 \citep{simonoff2003analyzing}. Requires attribution, and data are restricted to be used for scientific, educational and/or noncommercial purposes. License: Public
        \item Data set 41138 \citep{Dua:2019}. License: GNU GPL v3
        \item Data set 40978 \citep{Dua:2019,kushmerick1999learning}. License: Public
        \item Data set 40981 \citep{Dua:2019,quinlan1987simplifying}. License: Public
    \end{itemize}
    \item \simoptbench{}, based on SimOpt \citep{Eckman2023SimOptAT} (MIT license):
    \begin{itemize}[noitemsep,nolistsep]
        \item No underlying data set. We generate the changing utilities using a random walk.
    \end{itemize}
    \item XGBoost \citep{Chen_2016_XGBoost} (Apache License 2.0)
    \begin{itemize}[noitemsep,nolistsep]
        \item Data set: mnist\_784 downloaded from \citep{OpenML2013}
    \end{itemize}
\end{itemize}

The only data we handled was the MNIST data. It does not contain any personally identifiable or offensive content.

\section{XGBoost evaluations collection } \label{app-xgboost-additional}

For our XGBoost benchmark we collected evaluations of 1000 hyperparameter configurations on 28 training data set sizes, which we used as the basis of our simulation. 
The hyperparameter configurations were randomly selected and evaluated on each of the 28 data set sizes. 
The search space for the hyperparameters is given in \cref{tab-xgboost-hyperparameters} and the data set sizes in \cref{tab-xgboost-data-sizes}. 

\begin{table}[h]
\centering
\begin{tabular}{@{}lllrr@{}}
\toprule
Hyperparameter     & Type  & Min  & Max & Scaling \\ \midrule
learning\_rate
& Cont. & 1e-6 & 1   & log     \\
min\_child\_weight
& Cont. & 1e-6 & 32  & log     \\
max\_depth
& Int   & 2    & 32  & log     \\
n\_estimators
& Int   & 2    & 256 & log     \\ \bottomrule
\end{tabular}
\caption{Hyperparameter search space used for XGBoost.}
\label{tab-xgboost-hyperparameters}
\end{table}

\begin{table}[h]
\centering
\begin{tabular}{@{}rrrrrrrrr@{}}
\toprule
56     & 72   & 93   & 120   & 155   & 201   & 259   & 335   & 433   \\ 
560    & 723  & 934  & 1206  & 1558  & 2012  & 2599  & 3357  & 4335  \\
5600   & 7232 & 9341 & 12064 & 15582 & 20125 & 25992 & 33571 & 43358 \\
56000  &     &     &     &     &     &     &     &     \\ \bottomrule
\end{tabular}
\caption{Data set sizes evaluated for XGBoost.}
\label{tab-xgboost-data-sizes}
\end{table}

\section{Additional details on the experimental setup} \label{app-bo-start}

\textbf{Seeds:}
We rerun each method on each benchmark 50 times. We use integers between 0 and 49 as seeds.

The transfer learning methods require evaluations from previous tasks. We use \bayesopt{} to collect these. For all methods except \boundingbox{} this was done for one task (task 1). For \boundingbox{} we do two tasks (tasks 1 and 2), as otherwise the method collapses to only trying the best evaluation from the first task on any of the future tasks. More specifically, we run \bayesopt{} until we have different optima for at least two tasks, as otherwise we also get the situation of the bounding box only containing one configuration.

\section{\studentbo{} implementation details} 

The first $N$ configurations are used to evaluate the top  configuration from each of the previous $N$ tasks. We do this in reverse order, i.e. at task $i$ we first evaluate the configuration from task $i-1$, then from task $i-2$ and so on until task $i-N$. 

There are several situations which require modifications to this:
\begin{itemize}[noitemsep,nolistsep]
    \item There are $L<N$ previous tasks: in this case, we pick the top configuration from each of the $L$ tasks, and then pick the second-best configuration from each of the tasks, until we reach $N$ configurations. If we still don't have $N$ configurations we continue with the third-best, and so on.
    \item There are repeated optimal configurations: in this case the configuration is skipped. That means that we might end up using the top configuration from task $i-(N+1)$.
    \item There are joint optima: this is the situation if two hyperparameter configurations both perform best. We attempt to pick the first of these configurations, but if it is a repeat it will be skipped. We then add the other joint optima to the back of the list we are considering. So if there are $L<N$ previous tasks it might get selected once we have included one optima from each of the $L$ tasks. 
\end{itemize}

The code for \studentbo{} is available together with the rest of the paper code.

\section{BO details} \label{app-bo-details}

We use a slightly updated version of the BoTorch-based \citep{balandat2020botorch} BO in Syne Tune \citep{salinas2022syne}, see \url{https://github.com/sighellan/syne-tune/blob/othpo-results/syne_tune/optimizer/schedulers/searchers/botorch/botorch_searcher.py}. The acquisition function is a Monte Carlo version of expected improvement \citep{jones1998efficient}, and the covariance function Mat\'ern 5/2. We also apply input warping.

For \botransfer{} the maximum number of observations is set to 200. For later tasks, we subsample 200 of the past/current samples.

\section{Additional hyperparameter landscapes}

\subsection{XGBoost}

\begin{figure}[h]
    \centering
    \includegraphics[width=0.5\linewidth]{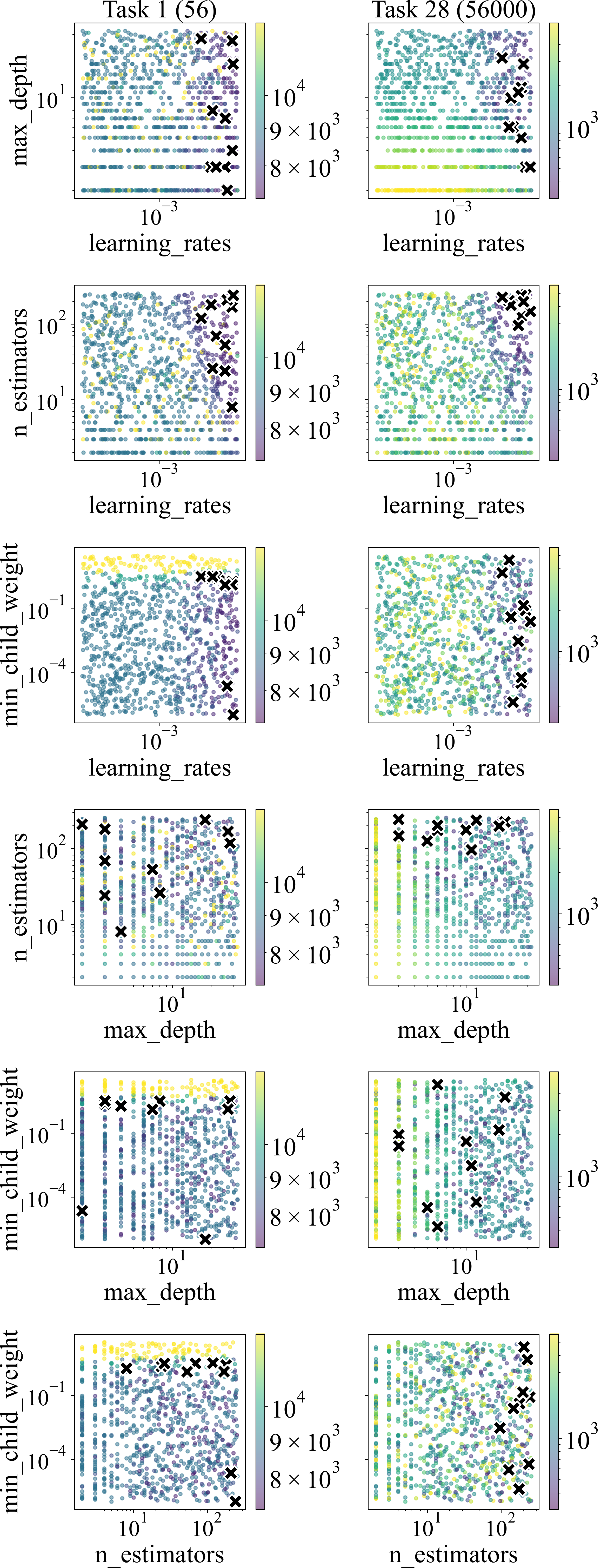}
    \caption{ Hyperparameter landscape plots for all combinations of two hyperparameters of the XGBoost benchmark. Left: Task 1 with 56 data points in the training set. Right: Task 28 with 56000 data points. Black crosses indicate the top 10 configurations.}
  \label{fig:xgboost-additional-landscapes}
\end{figure}

\cref{fig:xgboost-additional-landscapes} compares the top hyperparameters for the first and last tasks of the XGBoost benchmark. We show all six possible 2-dimensional combinations of the four hyperparameters.

\subsection{YAHPO} \label{additional-yahpo-plots}

\begin{figure}[h]
     \centering
    \begin{subfigure}[b]{\linewidth}
         \centering
         \includegraphics[width=\linewidth]{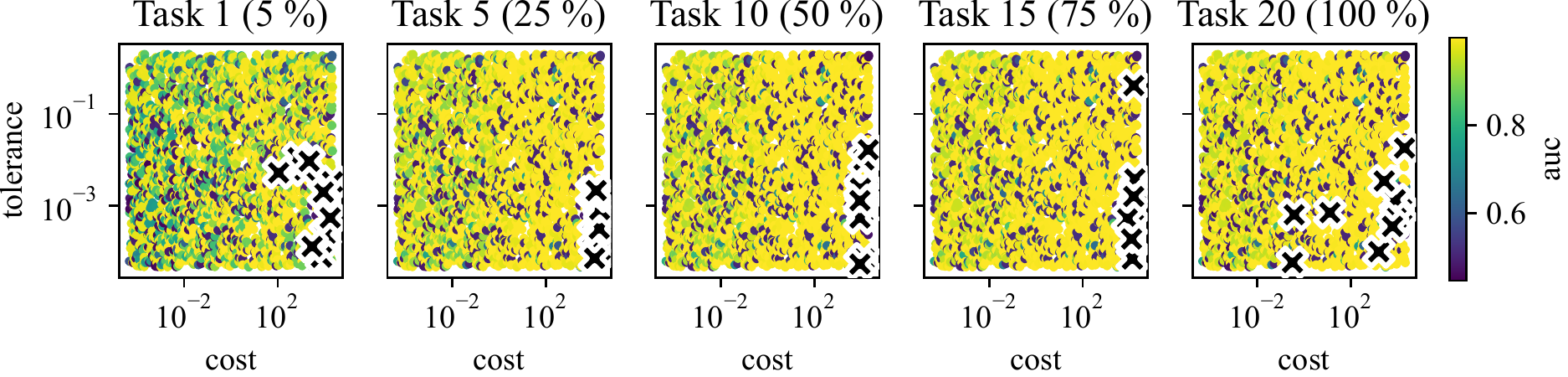}
         \caption{SVM 458}
     \end{subfigure}
     \begin{subfigure}[b]{\linewidth}
         \centering
         \includegraphics[width=\linewidth]{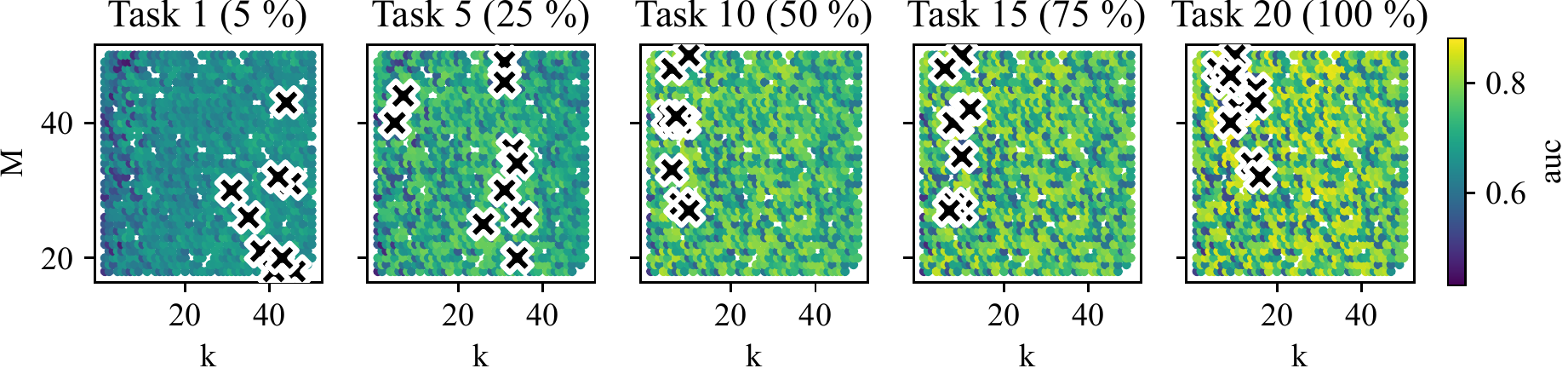}
         \caption{AKNN 4538}
     \end{subfigure}
     \begin{subfigure}[b]{\linewidth}
         \centering
         \includegraphics[width=\linewidth]{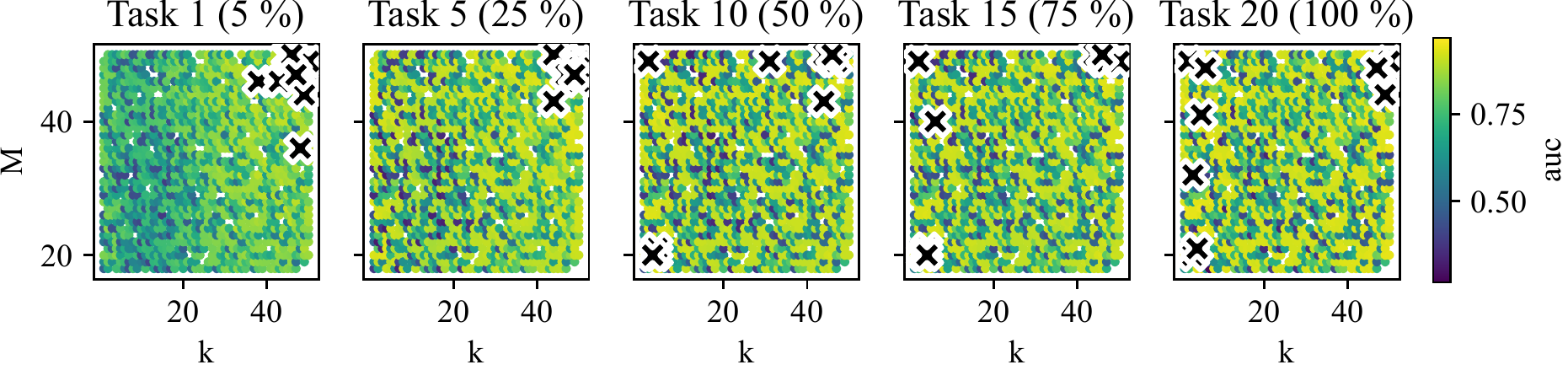}
         \caption{AKNN 41138}
     \end{subfigure}
    \caption{Further plots of YAHPO hyperparameter landscapes showing ordered behaviour on SVM and AKNN models.}
    \label{yahpo-additional-landscapes-1}
\end{figure}

\begin{figure}[h]
     \centering
     \begin{subfigure}[b]{\linewidth}
         \centering
         \includegraphics[width=\linewidth]{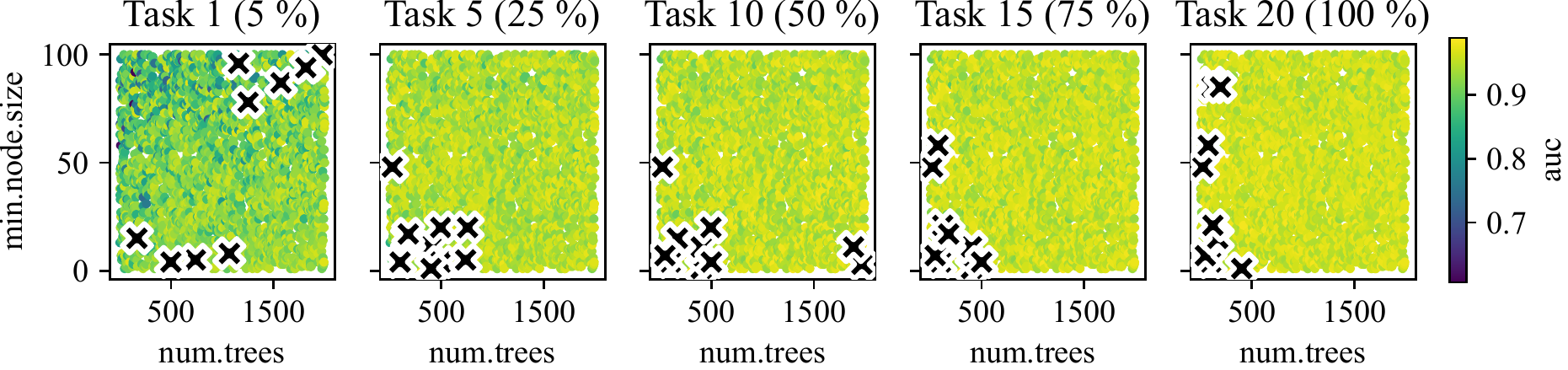}
         \caption{ranger 4154}
     \end{subfigure}
     \begin{subfigure}[b]{\linewidth}
         \centering
         \includegraphics[width=\linewidth]{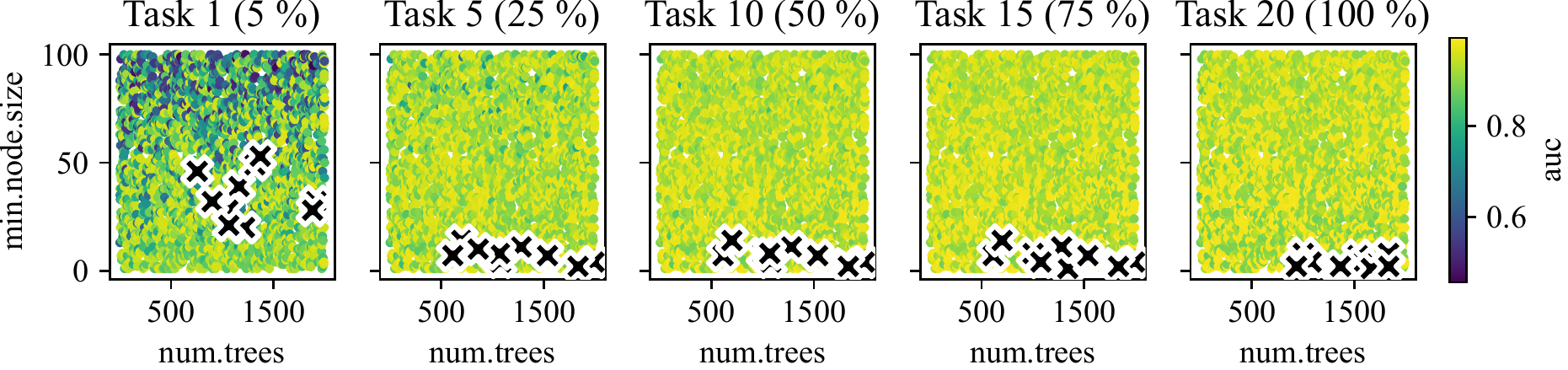}
         \caption{ranger 40978}
     \end{subfigure}
     \begin{subfigure}[b]{\linewidth}
         \centering
         \includegraphics[width=\linewidth]{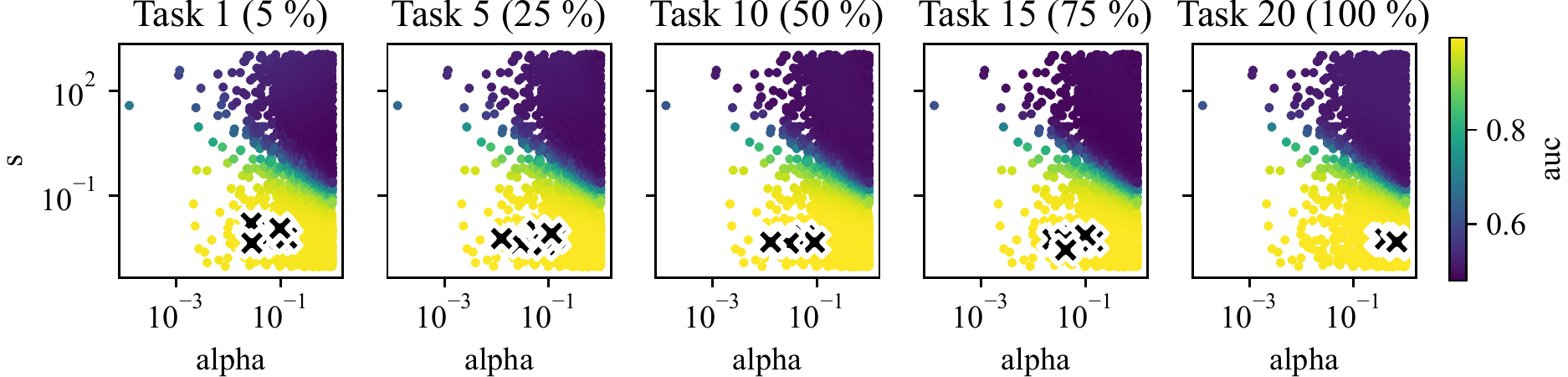}
         \caption{glmnet 375}
     \end{subfigure}
     \begin{subfigure}[b]{\linewidth}
         \centering
         \includegraphics[width=\linewidth]{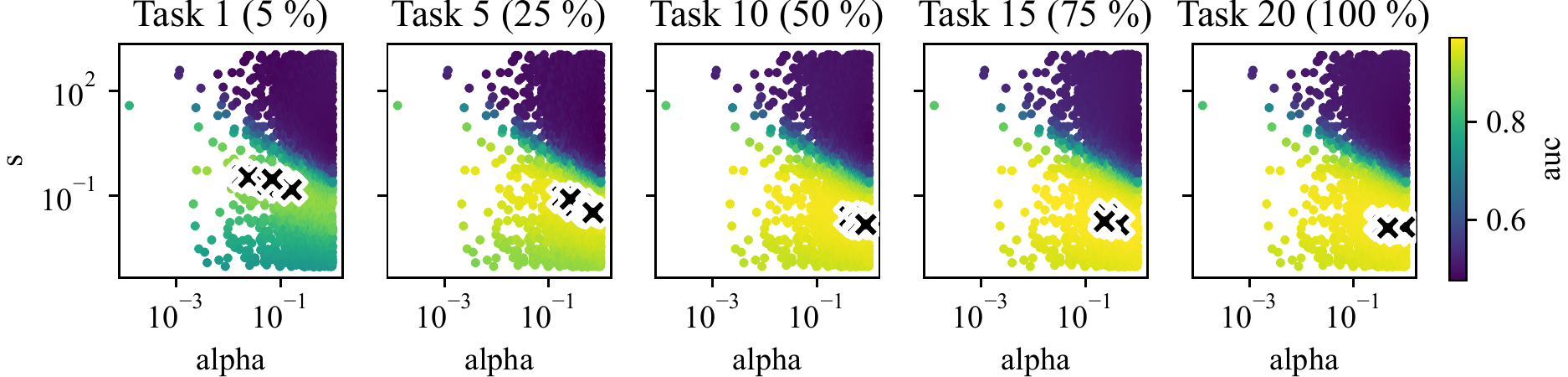}
         \caption{glmnet 40981}
     \end{subfigure}
        \caption{Further plots of YAHPO hyperparameter landscapes showing ordered behaviour on ranger and glmnet models.}
        \label{yahpo-additional-landscapes-2}
\end{figure}

\cref{yahpo-additional-landscapes-1,yahpo-additional-landscapes-2} show additional YAHPO hyperparameter landscapes for the model -- data set combinations used in \cref{fig:yahpo-additional-scenarios}. The final hyperparameter landscape used is shown in \cref{fig:yahpo-simopt-landscapes} (top).

\section{Additional results} \label{app-additional-results}

\subsection{Ablations} \label{app-ablations}

\begin{figure}[h]
    \centering
    \includegraphics[width=0.7\linewidth]{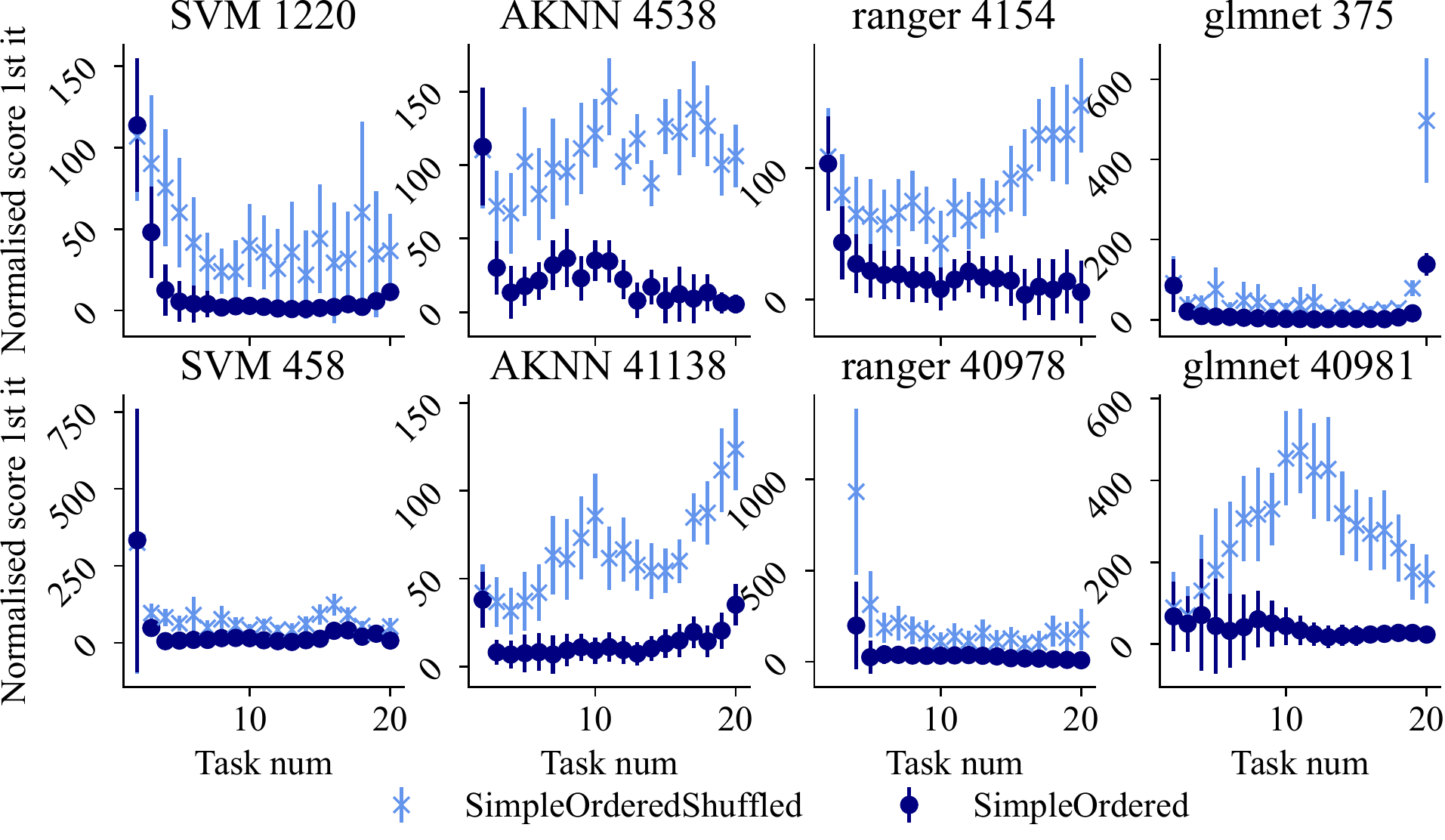}
    \caption{Mean normalised scores after first configuration evaluation for all eight YAHPO combinations considered.  We see that for almost all the tasks \studentbo{} performs better than its ablation \studentboshuffled{}.}
  \label{fig:app-ablations-warmboshuffled}
\end{figure}

We perform two ablation experiments. In \cref{fig:app-ablations-warmboshuffled} we compare \studentbo{} to \studentboshuffled{}, the version that takes top points from randomly chosen previous tasks. As can be seen, the ordered version does much better. 

\begin{figure}[h]
    \centering
    \includegraphics[width=0.7\linewidth]{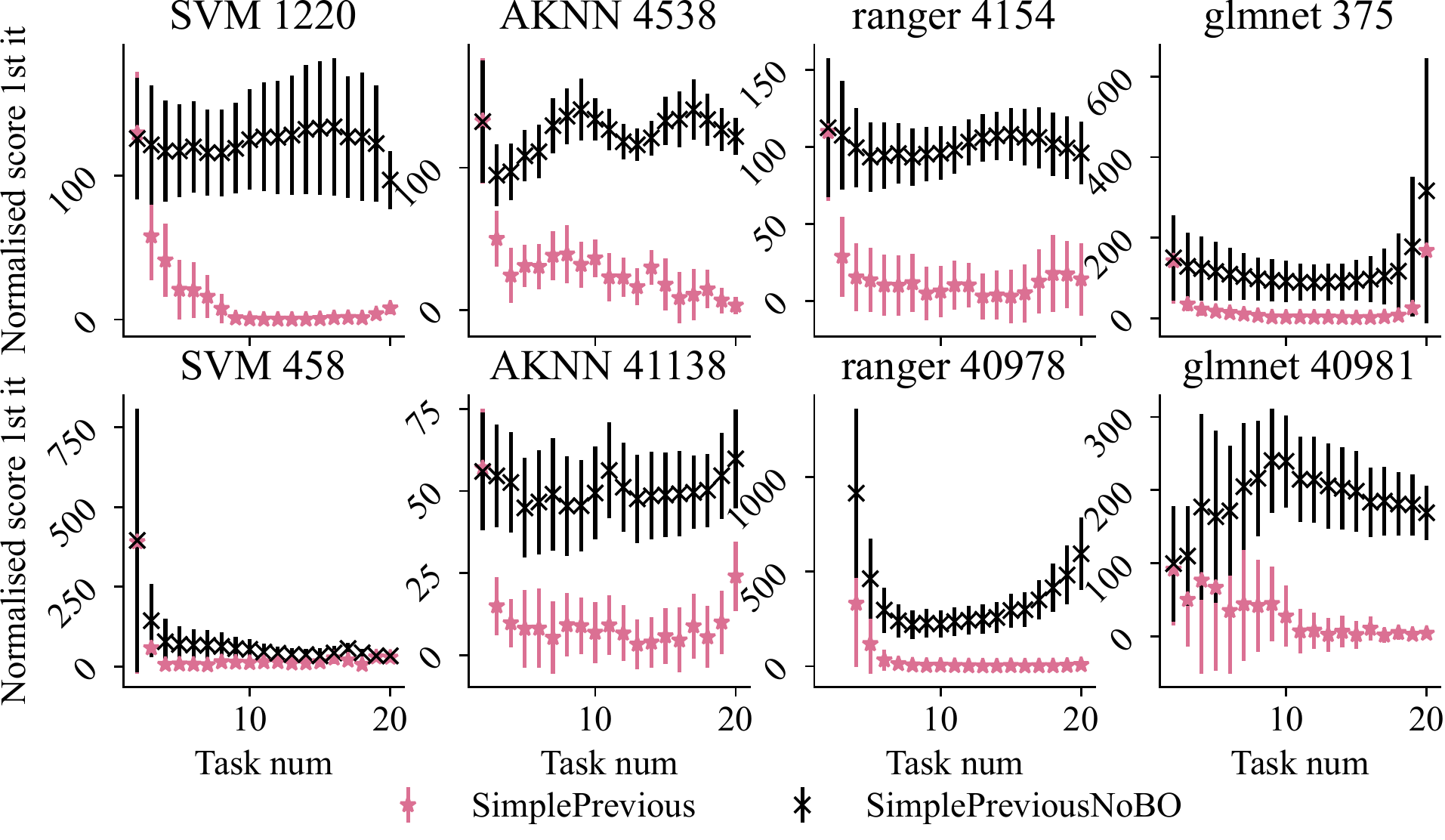}
    \caption{Mean normalised scores after first configuration evaluation for all eight YAHPO combinations considered.  We see that for almost all the tasks \prevbo{} performs better than its ablation \prevnobo{}.}
  \label{fig:app-ablations-prevnobo}
\end{figure}

In \cref{fig:app-ablations-prevnobo} we compare \prevbo{} to \prevnobo{}, the version that only considers configurations used in the previous task. That means that throughout the tasks only the configurations used by \bayesopt{} in the initial task are considered. As can be seen, the version that includes exploration through \bayesopt{} does much better. 

\subsection{Rankings over evaluations} \label{app-rankings}

\begin{figure}[h]
    \centering
    \includegraphics[width=0.8\linewidth]{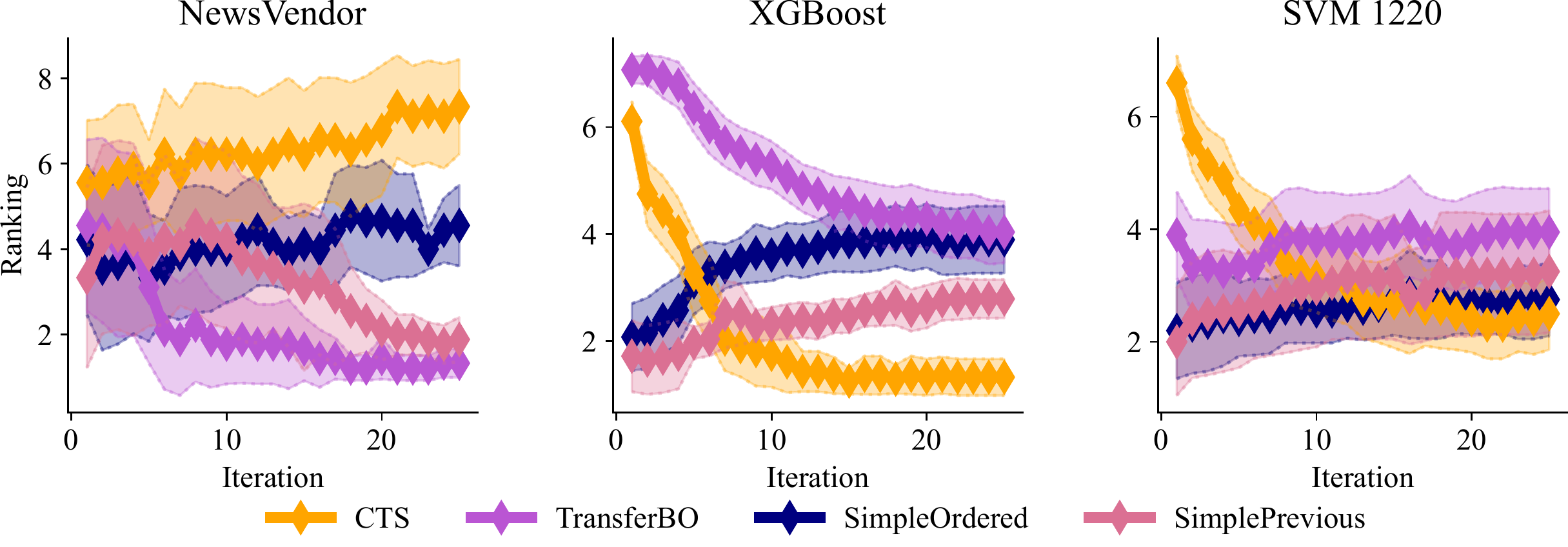}
    \caption{Mean rankings ($\pm$ 2 standard error) as a function of configurations evaluated, averaged over tasks.}
  \label{fig:app-rankings}
\end{figure}

\cref{fig:app-rankings} shows the mean internal rankings among the ten methods for a subset of the methods. We see that \studentbo{} and \prevbo{} do very well at the beginning, and are then overtaken by either \botransfer{} or \quantiles{}. 

\subsection{Additional bar plots } \label{app-additional-bars}

\begin{figure}[h]
    \centering
    \includegraphics[width=0.85\linewidth]{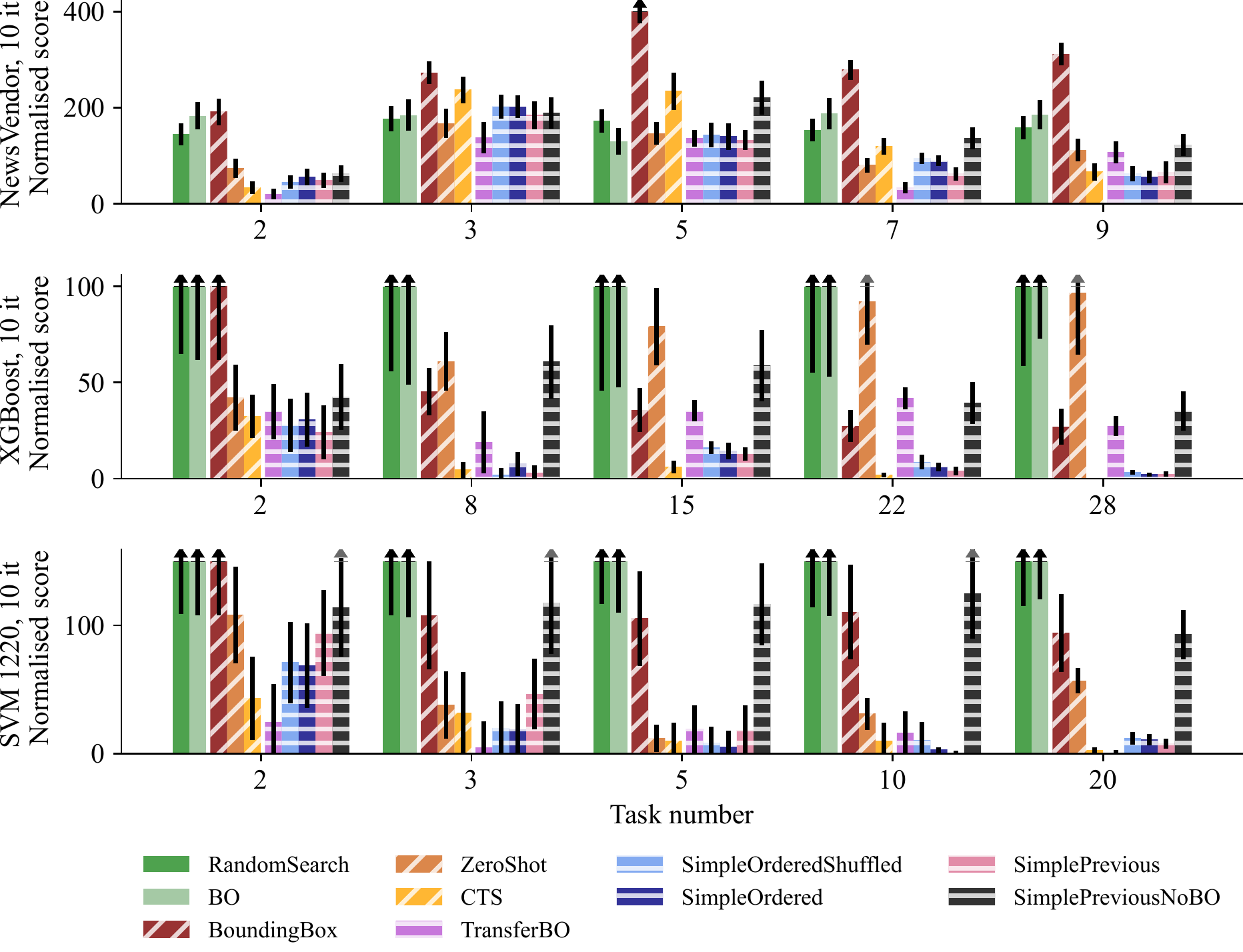}
    \caption{Mean normalised scores (+- 2 standard error) after the tenth configuration (lower is better). \simoptbench{} (top), XGBoost (middle) and \svmfirst{} (bottom).  Black arrows indicate that the mean was above the plotted range, grey arrows that the standard error range was above. Version of \cref{fig:all-normalised-task} for the tenth configuration.}
  \label{fig:all-normalised-task-10}
\end{figure}

\begin{figure}[h]
    \centering
    \includegraphics[width=0.85\linewidth]{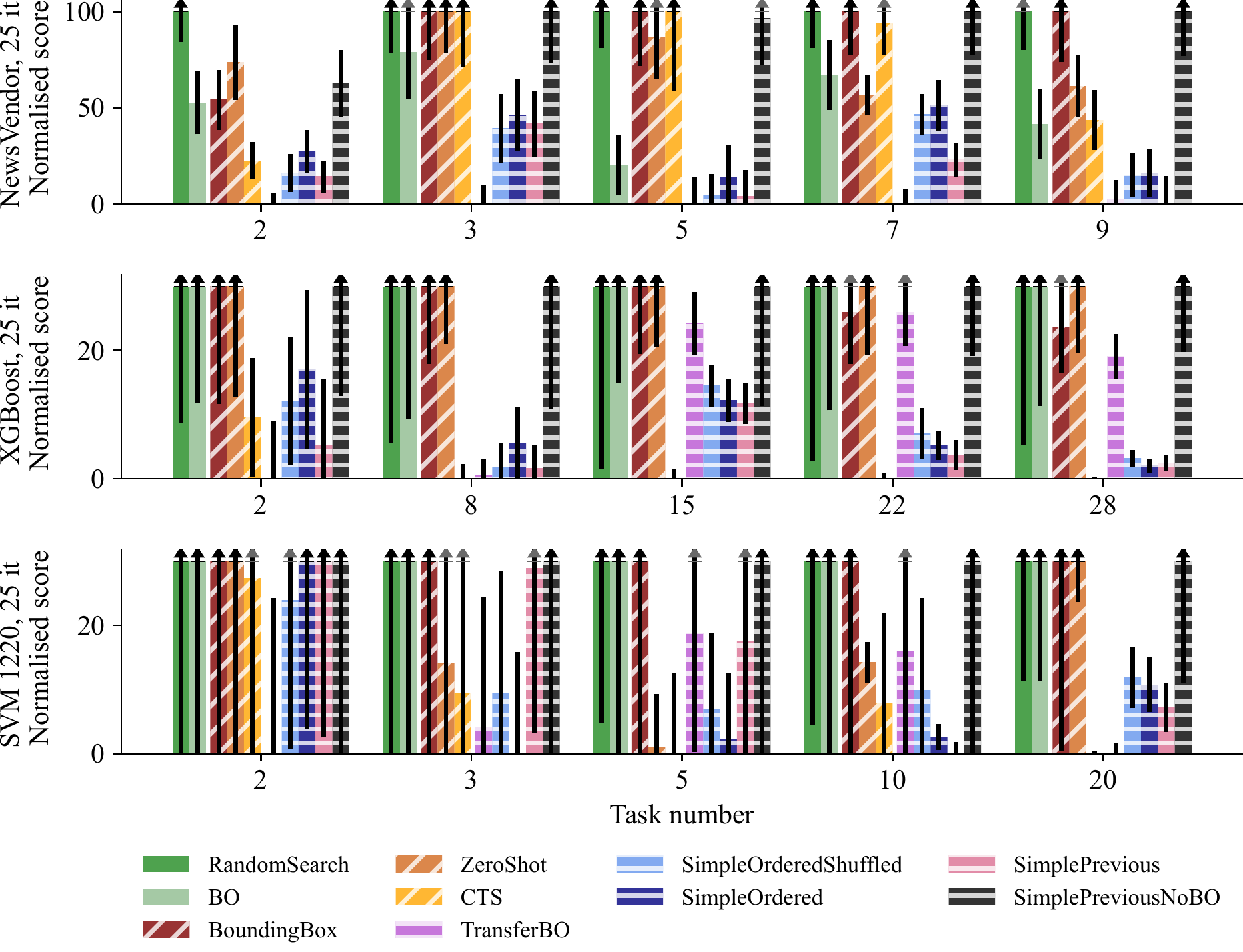}
    \caption{Mean normalised scores (+- 2 standard error) after the 25th configuration (lower is better). \simoptbench{} (top), XGBoost (middle) and \svmfirst{} (bottom).  Black arrows indicate that the mean was above the plotted range, grey arrows that the standard error range was above. Version of \cref{fig:all-normalised-task} for the 25th configuration.}
  \label{fig:all-normalised-task-25}
\end{figure}

\cref{fig:all-normalised-task-10,fig:all-normalised-task-25} are versions of \cref{fig:all-normalised-task} for 10 and 25 configuration evaluations, respectively.

\subsection{Optimisation curves} \label{app-additional-iteration-curves}

\begin{figure}[h]
     \centering
     \begin{subfigure}[b]{\linewidth}
         \centering
         \includegraphics[width=\linewidth]{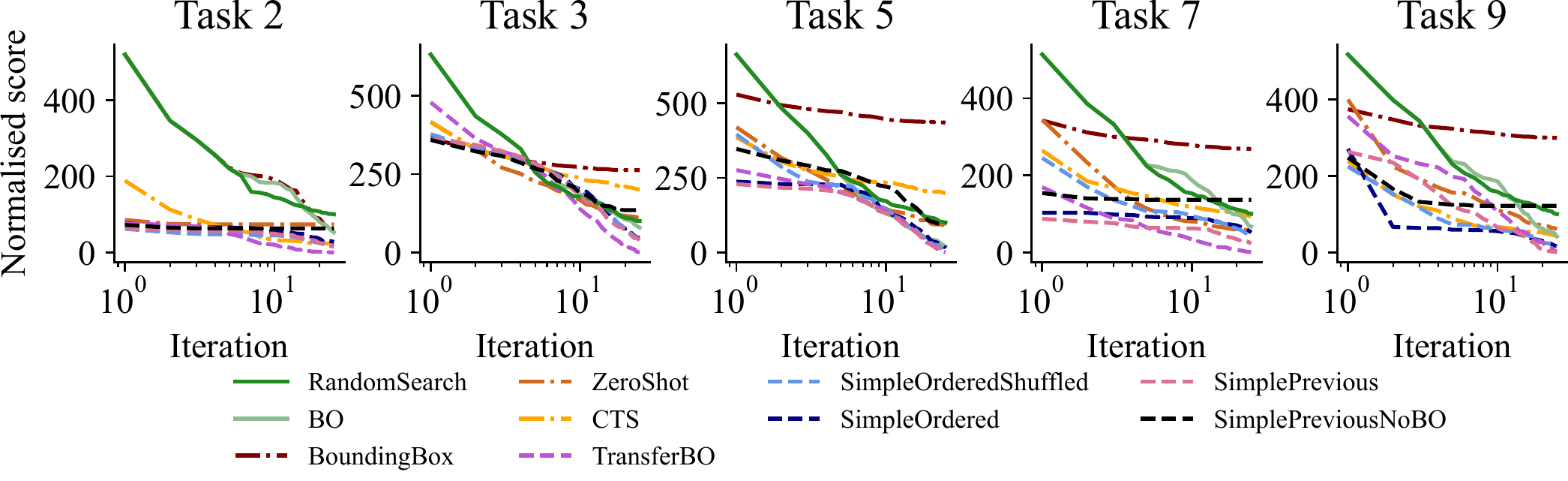}
         \caption{SimOpt}
     \end{subfigure}
     \begin{subfigure}[b]{\linewidth}
         \centering
         \includegraphics[width=\linewidth]{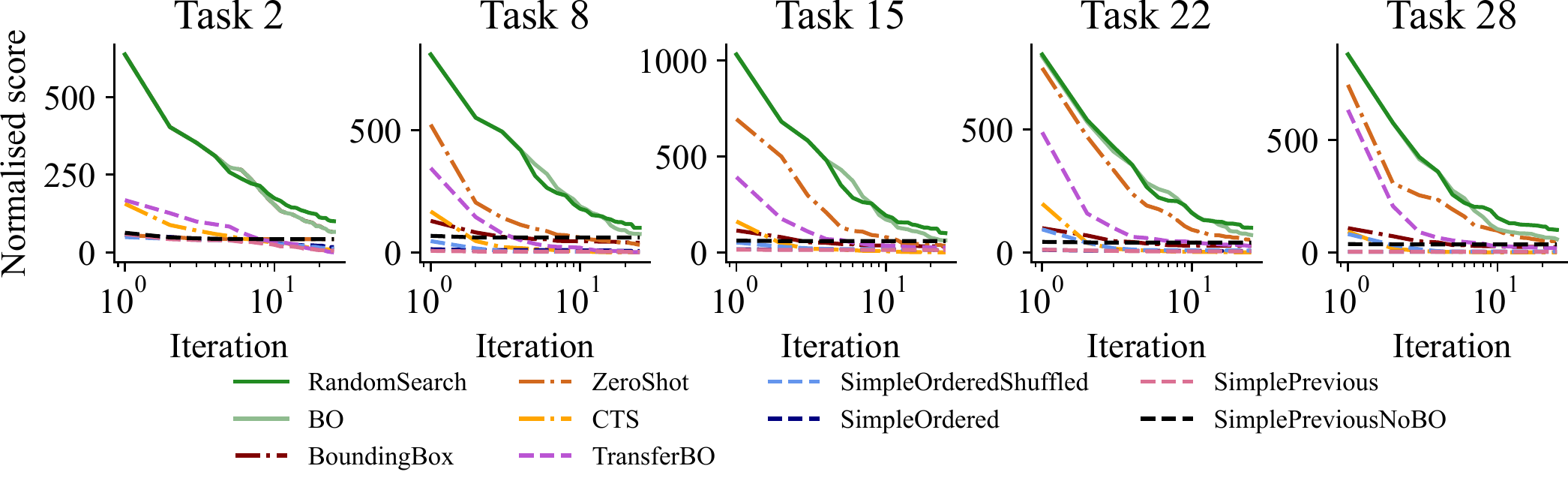}
         \caption{XGBoost}
     \end{subfigure}
     \begin{subfigure}[b]{\linewidth}
         \centering
         \includegraphics[width=\linewidth]{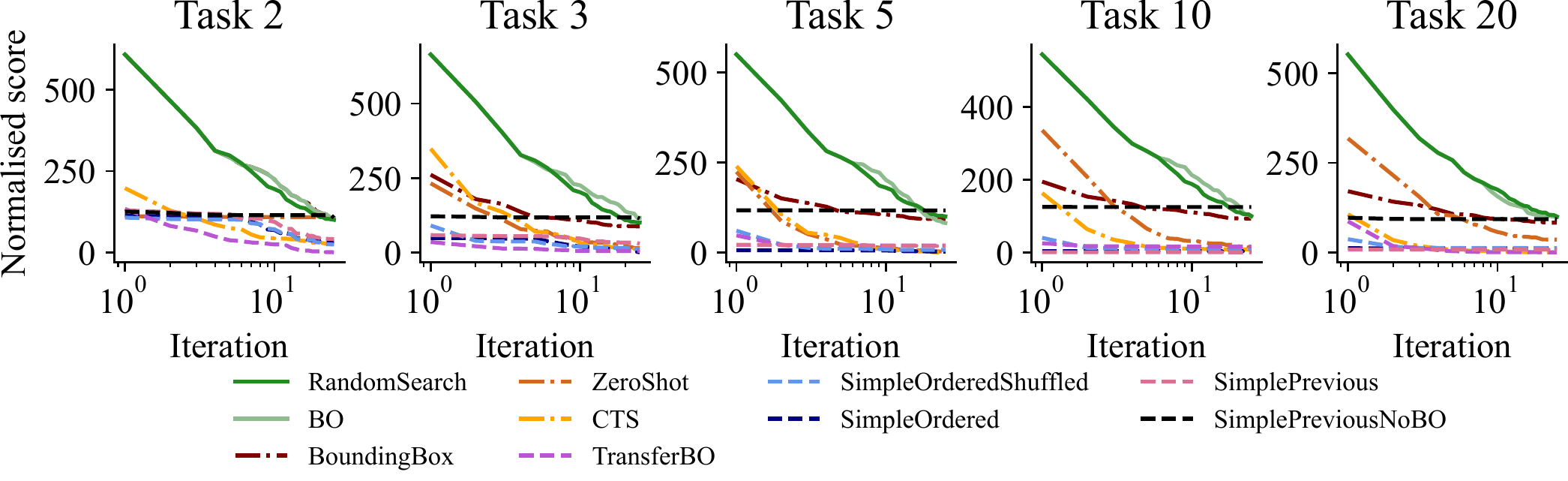}
         \caption{\svmfirst{}}
     \end{subfigure}
        \caption{Companion plot to \cref{fig:all-normalised-task}, showing the normalised score as a function of configurations evaluated for the same subset of tasks. \studentbo{} starts off well, but does not improve much with more iterations.}
        \label{fig:iteration-curves}
\end{figure}

We show the optimisation curves for \simoptbench{}, XGBoost and \svmfirst{} in \cref{fig:iteration-curves}.

\subsection{Downstream performance} \label{app-downstream}

\cref{tab:downstream-comparison} shows that the downstream performance of \studentbo{} is better than \quantiles{}. The reduction in standard error is calculated as $100(1-s^{(i)}_{\mathrm{SO}}/s^{(i)}_{\mathrm{\quantiles{}}})$ for each task $i$ where $s^{(i)}_{\mathrm{\quantiles{}}}$ is the standard error of the performance across replications for \quantiles{}. $s^{(i)}_{\mathrm{SO}}$ is the same for \studentbo{}. 

The improvement in mean is calculated as $100(1-m^{(i)}_{\mathrm{SO}}/m^{(i)}_{\mathrm{\quantiles{}}})$ for maximisation, and the same multiplied by -1 for minimisation. Here $m^{(i)}_{\mathrm{SO}}$ and $m^{(i)}_{\mathrm{\quantiles{}}}$ are the means across replications for \studentbo{} and \quantiles{}, respectively.
Note that we do not include the between-seed variation in our error estimate of the improvement in mean, only the between-task variation. 

\begin{table}[]
\centering
\begin{tabular}{@{}lrcrcrc@{}}
\toprule
                                                                                & \multicolumn{2}{l}{\simoptbench{}} & \multicolumn{2}{l}{XGBoost} & \multicolumn{2}{l}{\svmfirst{}} \\ \midrule
\begin{tabular}[c]{@{}l@{}}Reduction in standard error\\ \studentbo{} over  \quantiles{} (\%) \end{tabular} & 61.3      & (53.7 -- 68.9)      & 92.5    & (88.9 -- 96.0)    & 89.4     & (83.0 -- 95.9)    \\
\begin{tabular}[c]{@{}l@{}}Improvement in mean \\ \studentbo{} over  \quantiles{} (\%) \end{tabular}                                                               & 21.7     & (8.0 -- 35.4)     & 22.5   & (18.2 -- 26.7)  & 5.8     & (4.9 -- 6.7)    \\ \bottomrule
\end{tabular}
\caption{Downstream comparison of \studentbo{} and \quantiles{} after one configuration. \studentbo{} gives an improvement on all benchmarks, and the standard error is smaller, making it more reliable. We present means +- two standard errors for each value.}
\label{tab:downstream-comparison}

\end{table}

\subsection{Sampling locations} \label{app-sampling-locations}

\begin{figure}[h]
    \centering
    \includegraphics[width=0.95\linewidth]{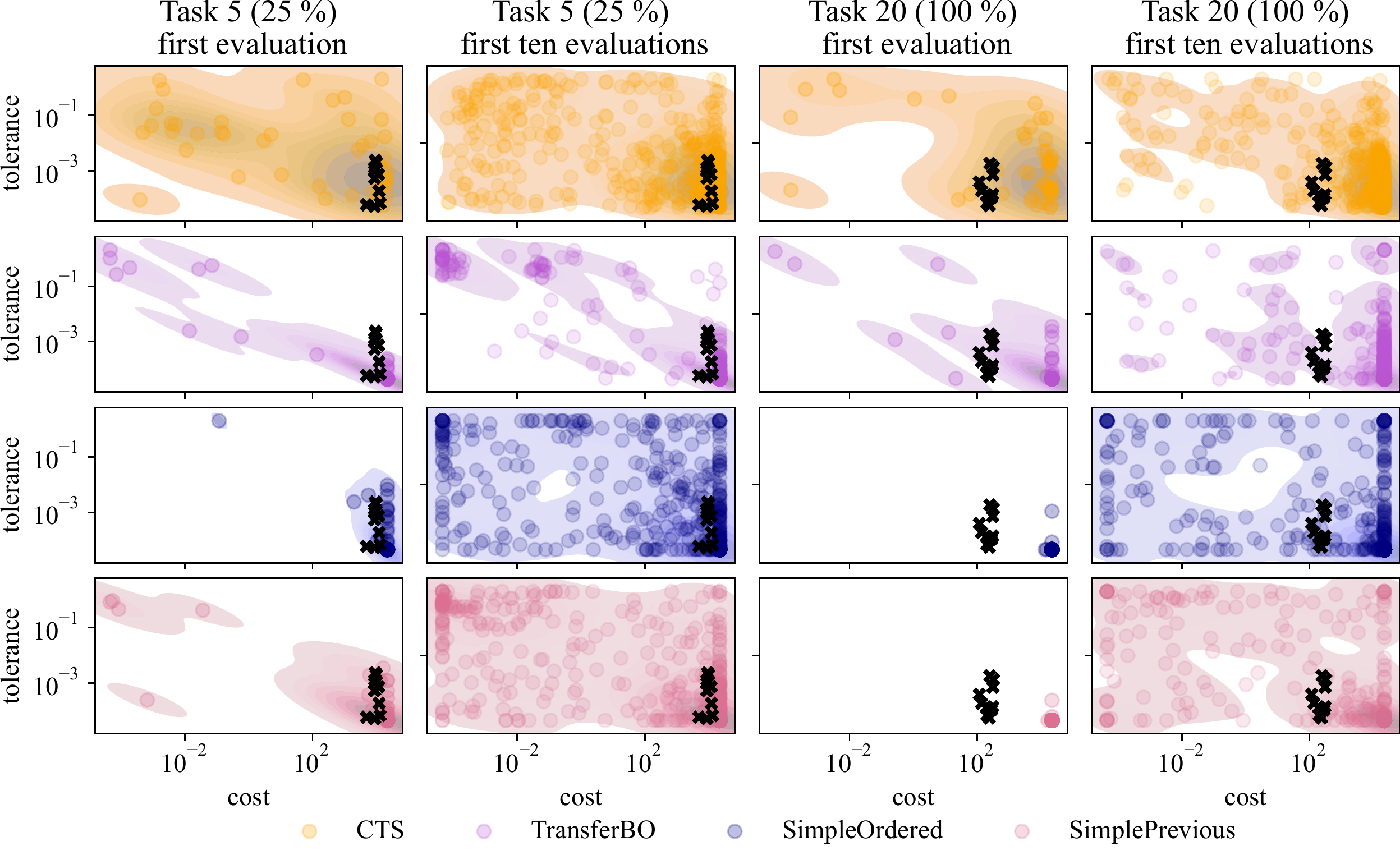}
    \caption{Sampling locations of the first and first ten evaluations for YAHPO \svmfirst{} across the fifty replications. Black crosses indicate the top ten hyperparameter configurations. As can be seen, the \studentbo{} and \prevbo{} samples are much more concentrated than the others for the first evaluation, and then is very explorative once the first five evaluations are done.  }
  \label{fig:sampling-locations}
\end{figure}

\cref{fig:sampling-locations} shows sampling locations. As can be seen, \studentbo{} and \prevbo{} combine very focused early exploitation with broad exploration later on. 

\subsection{Ordering with different surrogate model: Density-Ratio Estimation}
\label{app-dre-model}

This section presents ablation results of combining the ordered approach in \studentbo{} with BO using density-ratio estimation for the surrogate model \citep{tiao2021bore}: \dreordered{}. Note that the figures in the main paper were not replotted with these new results, although the metrics of the other methods can be impacted by the introduction of a new method. We also did not update the values in \cref{app-compute-budget} to include these extra experiments.

\begin{figure}[h]
    \centering
    \includegraphics[width=0.85\linewidth]{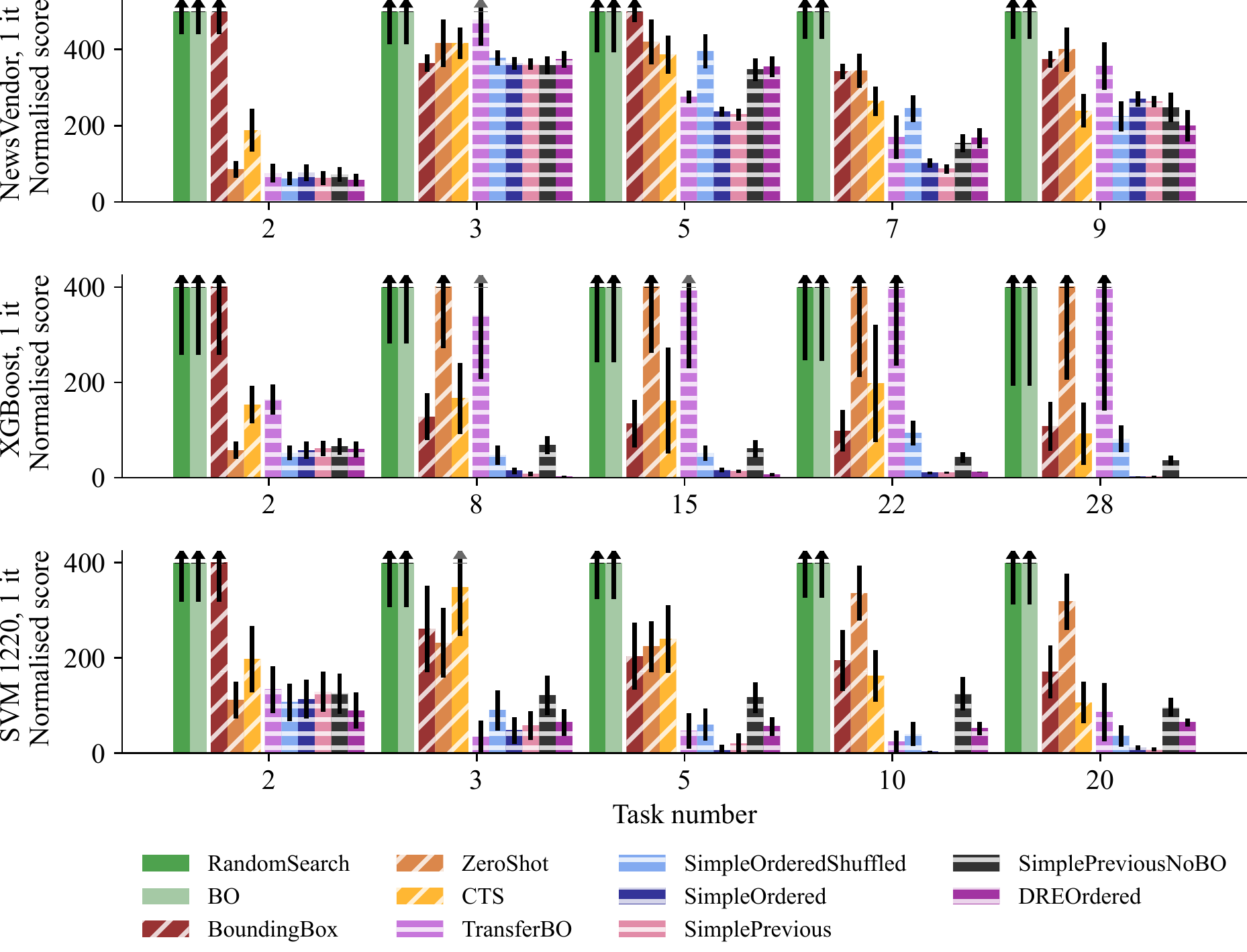}
    \caption{Mean normalised scores (+- 2 standard error) after the first configuration (lower is better). \simoptbench{} (top), XGBoost (middle) and \svmfirst{} (bottom).  Black arrows indicate that the mean was above the plotted range, grey arrows that the standard error range was above. Version of \cref{fig:all-normalised-task} with the addition of \dreordered{} as a baseline.}
  \label{fig:all-normalised-task-1_DREOrdered}
\end{figure}

\begin{figure}[h]
    \centering
    \includegraphics[width=0.85\linewidth]{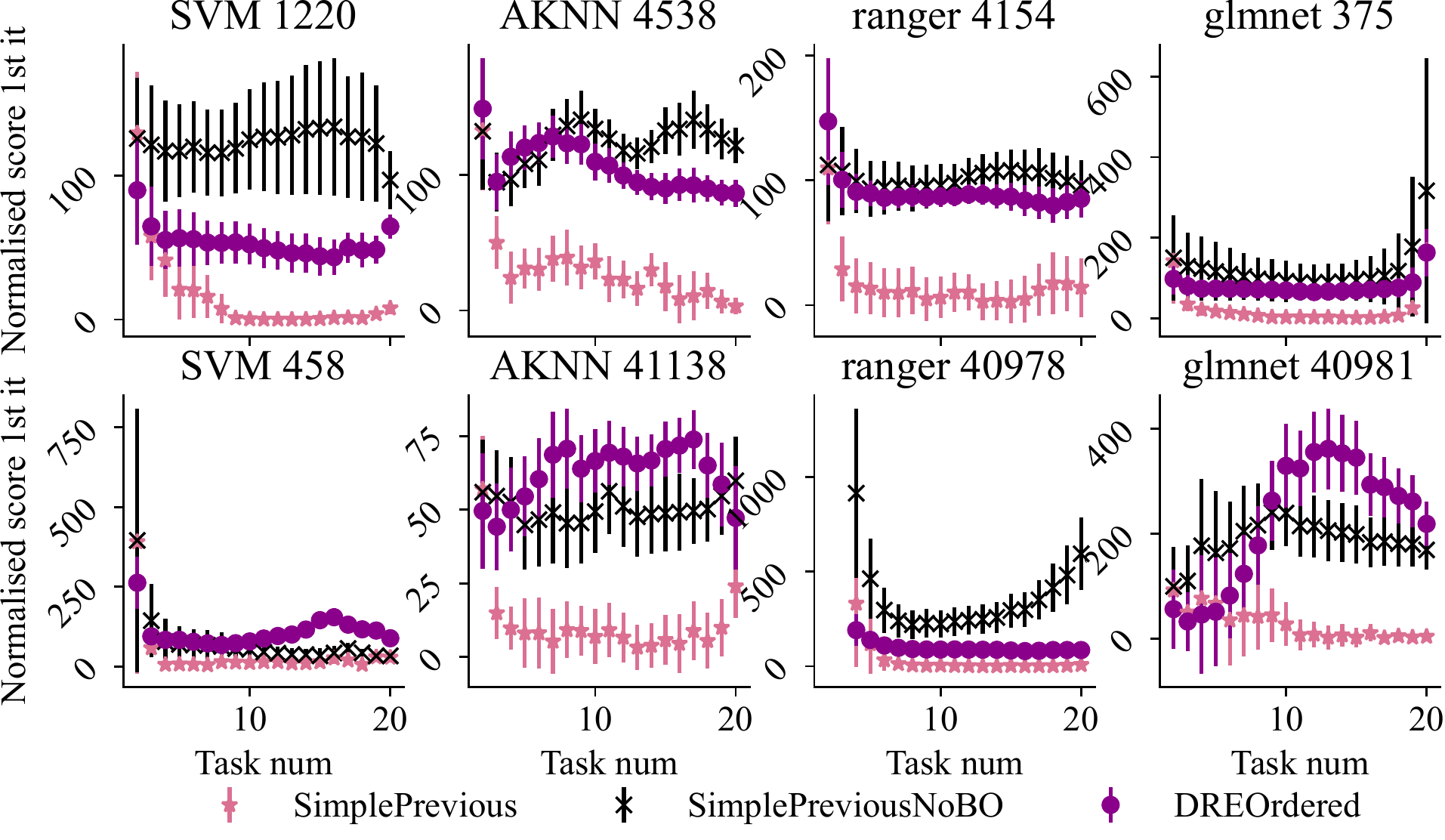}
    \caption{Comparison of new baseline \dreordered{} with \prevbo{} and \prevnobo{}.}
  \label{fig:DREOrdered_YAHPO}
\end{figure}

We present results for \dreordered{} in \cref{fig:all-normalised-task-1_DREOrdered,fig:DREOrdered_YAHPO}.
As can be seen, the performance of \dreordered{} is between that of \prevbo{} and \prevnobo{}. This suggests that Gaussian processes are better surrogate models for our problem. But comparing \dreordered{} to \stantrans{} we see that also with this surrogate model we outperform the non-ordered methods.

\section{Compute budget} \label{app-compute-budget}

This summaries the compute costs for the results included in the paper (not including \cref{app-dre-model}). 

The experiments were run on AWS Sagemaker, using \lstinline{ml.c5.18xlarge} compute instances, which have 72 vCPUs, and 144 GiB memory. We ran a total of 65 experiments: 10 for \simoptbench{}, 10 for XGBoost, 10 for SVM 1220 (YAHPO) and 5 for each of the 7 remaining YAHPO combinations, so 35. We also trained and evaluated a total of 28000 XGBoost models for our XGBoost benchmark, also on AWS Sagemaker.

Collecting the XGBoost evaluations took a total of 79 hours and 46 minutes of compute time.

Collecting the experiment evaluations took a total of 231 hours and 39 minutes. 

In total, we used 311 hours and 26 minutes of compute time for the results presented.

\end{document}